\documentclass{article}

\usepackage{microtype}
\usepackage{graphicx}
\usepackage{booktabs} % for professional tables
\usepackage{enumitem} 
\usepackage{subcaption}
\usepackage{multirow}
\usepackage{floatrow}
\usepackage{color}
\usepackage{colortbl}
\newfloatcommand{capbtabbox}{table}[][\FBwidth]

\usepackage{hyperref}

\usepackage[accepted]{icml2024}

\usepackage{amsmath}
\usepackage{amssymb}
\usepackage{mathtools}
\usepackage{amsthm}
\usepackage{amsmath,amsfonts,bm}

\def\eqref#1{equation~\ref{#1}}
\def\1{\bm{1}}

\def\rvx{{\mathbf{x}}}

\DeclareMathAlphabet{\mathsfit}{\encodingdefault}{\sfdefault}{m}{sl}
\SetMathAlphabet{\mathsfit}{bold}{\encodingdefault}{\sfdefault}{bx}{n}

\usepackage[capitalize,noabbrev]{cleveref}

\theoremstyle{plain}

\theoremstyle{definition}

\theoremstyle{remark}

\usepackage[textsize=tiny]{todonotes}

\icmltitlerunning{Revealing the Dark Secrets of Extremely Large Kernel ConvNets on Robustness}

\begin{document}

\twocolumn[
\icmltitle{Revealing the Dark Secrets of Extremely Large Kernel ConvNets on Robustness}



\icmlsetsymbol{equal}{*}

\begin{icmlauthorlist}
\icmlauthor{Honghao Chen}{yyy,comp}
\icmlauthor{Yurong Zhang}{stju}
\icmlauthor{Xiaokun Feng}{yyy,comp}
\icmlauthor{Xiangxiang Chu}{sch}
\icmlauthor{Kaiqi Huang}{yyy,comp}
\end{icmlauthorlist}

\icmlaffiliation{yyy}{CRISE, Institute of Automation, Chinese Academy of Sciences
}
\icmlaffiliation{comp}{School of Artificial Intelligence, University of Chinese Academy of Sciences}
\icmlaffiliation{sch}{Meituan. Work done during the first two authors' internship at Meituan}
\icmlaffiliation{stju}{Shanghai Jiao Tong University}

\icmlcorrespondingauthor{Kaiqi Huang}{kaiqi.huang@nlpr.ia.ac.cn}
%\icmlcorrespondingauthor{Firstname2 Lastname2}{first2.last2@www.uk}

\icmlkeywords{Machine Learning, ICML}

\vskip 0.3in
]



\printAffiliationsAndNotice{}  % leave blank if no need to mention equal contribution

\begin{abstract}
Robustness is a vital aspect to consider when deploying deep learning models into the wild. Numerous studies have been dedicated to the study of the robustness of vision transformers (ViTs), which have dominated as the mainstream backbone choice for vision tasks since the dawn of 2020s.  Recently, some large kernel convnets make a comeback with impressive performance and efficiency. However, it still remains unclear whether large kernel networks are robust and the attribution of their robustness. In this paper, we first conduct a comprehensive evaluation of large kernel convnets' robustness and their differences from typical small kernel counterparts and ViTs on six diverse robustness benchmark datasets. Then to analyze the underlying factors behind their strong robustness, we design experiments from both quantitative and qualitative perspectives to reveal large kernel convnets' intriguing properties that are completely different from typical convnets. 
Our experiments demonstrate for the first time that pure CNNs can achieve exceptional robustness comparable or even superior to that of ViTs. Our analysis on occlusion invariance, kernel attention patterns and frequency characteristics provide novel insights into the source of robustness. Code available at: \url{https://github.com/Lauch1ng/LKRobust}.
\end{abstract}

\section{Introduction}
\label{sec1}
\setlist[itemize]{itemsep=0pt}

Over the past decade, the evolution of deep learning has been tightly linked with Convolutional Neural Networks (CNNs)~\cite{lecun1998gradient,alexnet,vgg, resnet}, which have played a crucial role in propelling the advancement of artificial intelligence. Nonetheless, with the dawn of the 2020s, the advent of Vision Transformers (ViTs)~\cite{vit,swin,touvron2021training,pvt}  has profoundly shaken the prevailing hegemony of CNNs. Facilitated by the self-attention mechanism, ViTs have manifested state-of-the-art performance in a variety of vision tasks such as image classification~\cite{vit, swin}, object detection~\cite{detr, dynamic_detr, deformable_detr, conditional_detr}, semantic segmentation~\cite{maskformer, mask2former} and self-supervised learning~\cite{beit,mae,ddae}, demonstrating their exceptional properties as the choice of foundation models.

As a versatile and widely used backbone, a key attribute of ViT is its strong robustness.   Robustness is a vital aspect to consider when deploying deep learning models into the wild~\cite{paul2022vision}.   Specifically, for safety-critical applications such as autonomous vehicles, robots, and healthcare, the learned representations must be robust, and without strong robustness the models can not be practically implemented~\cite{naseer2021intriguing}. Numerous studies have been dedicated to the investigation on evaluating the robustness of ViTs~\cite{mahmood2021robustness,naseer2021intriguing,paul2022vision,shao2021adversarial,bhojanapalli2021understanding}. They demonstrate that ViTs possess universal and strong robustness, which excel on various robust datasets~\cite{imagenet_ao,imagenet_cp,imagenet_r,imagenet_9} and exhibit clear superiority over typical CNNs in dealing with particular disturbances, such as image occlusion and adversarial attack~\cite{naseer2021intriguing}. The strong robustness paves the way for ViTs' extensive application across diverse real-world scenarios.

On the other hand, concerted efforts are dedicated to the revival of CNNs.   Notably, recent advances~\cite{convnext,replk,slak,unireplknet,pelk} in revisiting large kernel designs have revealed that when equipped with large kernel size (e.g., $31\times31$), pure CNN architecture can perform on par with or even better than state-of-the-art ViTs.  Large kernel convolution can substantially increase the model's effective receptive field~\cite{replk}, enabling large kernel convnets to have remarkable performances across various vision tasks~\cite{convnext,convnext_v2,replk,slak,largekernel3d,link,lkteacher}, such as classification, segmentation and detection, etc. Nevertheless, it still remains unclear whether large kernel networks are robust and the attribution of their robustness, which are of vital importance and could significantly impact their practical application and development.  It is natural to ask: \textit{are large kernel networks inherently robust?  How do they differ in terms of robustness from typical CNNs and ViTs?} And if they are robust, \textit{what contributes to their robustness?} In this paper, we delve into the above questions, providing empirical evidence and novel insights to reason about the robustness of large kernel convnets.

To answer the first two questions, we conduct thorough experiments on six diverse and widely-used robustness benchmark datasets. These datasets comprehensively evaluate model robustness from multiple perspectives including natural adversarial~\cite{imagenet_ao}, common corruptions~\cite{imagenet_cp}, semantic shifts~\cite{imagenet_r}, out-of-domain distribution~\cite{imagenet_ao},  common perturbations~\cite{imagenet_cp} and background dependency~\cite{imagenet_9}. In stark contrast to traditional CNNs, large kernel networks exhibit exceptionally strong robustness, comparable or even superior to ViT. %For instance, 

To comprehend the underlying factors for the superior robustness of large kernel networks, we systematically designed nine experiments to provide insights into their robustness from both quantitative and qualitative perspectives. These experiments and visualizations unveil intriguing properties of large kernel convnets, such as occlusion invariance, kernel attention patterns and frequency characteristics, which help understand the reasons behind their robustness. Our contributions can be summarized as follows:
\begin{itemize}[leftmargin=20pt]
\item [$\bullet$] We explore the robustness of large kernel networks across six ImageNet datasets concerning different types of robustness evaluation and conduct a comprehensive comparison with typical CNNs and ViT. We demonstrate that large kernel convnets' robustness significantly differs from typical CNNs and is comparable or even superior to that of ViT.

\item[$\bullet$] We devise nine experiments to investigate the robustness of large kernel convnets from both quantitative and qualitative perspectives, spreading occlusion invariance, adversarial attack, model perturbations, frequency characteristics and kernel attention pattern, etc.

\item[$\bullet$]Our analysis provides insights into the robustness of large kernel convnets, promoting their application and broader development. Moreover, our study is the first to reveal that pure CNNs can achieve comparable or even superior robustness to ViT, indicating that self-attention is not the only route to strong robustness and making a further step to the revival of CNNs.

\end{itemize}

\section{Related Work}
\label{sec2}

\subsection{Large Kernel ConvNets}
Large kernel convolutional networks can be traced back to a few traditional models from the early stages of deep learning~\cite{alexnet,szegedy2015going,szegedy2016rethinking}. However, it was the rise of small convolutional networks, represented by VGG-Net~\cite{vgg} and ResNet~\cite{resnet}, that facilitated the success of CNNs. And ever since then, a stack of small kernels (e.g., $1\times1$ or $3\times3$) became the mainstream choice for convnet design. Large kernel convolution has received little attention and has even been found to be detrimental to Imagenet performance~\cite{gcn}. Recently, works represented by ConvNeXt~\cite{convnext} and RepLKNet~\cite{replk} have sparked a revival of large kernel convolutional networks~\cite{convmixer, convnext, convnext_v2, replk, slak, pelk}. They demonstrate that large kernels can effectively enhance model performance, especially in downstream tasks~\cite{replk}. Large kernel design has also been introduced into other fields, such as 3D backbone~\cite{largekernel3d, link}, continuous convolutions~\cite{ckconv,flexconv}, knowledge distillation~\cite{lkteacher} and multi-modal learning~\cite{unireplknet}. Despite these advances, what remains largely unexplored is their robustness evaluation and attribution, which could significantly impact their practical application and development. Different from previous works, this paper investigates the robustness of large kernel networks, designing experiments to analyze their intriguing properties from multiple perspectives and making a further step for large kernel's revival.

\subsection{Robustness for ViTs}
As a versatile backbone, ViTs~\cite{vit, swin} have demonstrated impressive performance across a variety of vision tasks~\cite{vit,swin, dynamic_detr, deformable_detr, conditional_detr,maskformer,mask2former}. A key advantage of ViT over CNN is its strong robustness, which paves the way for ViTs’ extensive application across diverse real-world scenarios. Many works seek to study the robustness of ViTs from different perspectives. Compared with ResNet~\cite{resnet}, Bhojanapalli \textit{et al}.~\cite{bhojanapalli2021understanding} studied improved robustness of ViTs when evaluated on adversarial and natural adversarial examples. Shao \textit{et al}.~\cite{shao2021adversarial}  revealed that ViTs have better adversarial robustness over CNNs, attributed largely to their ability to learn highly generalizable high-frequency features, while convolutional layers appear to hinder. Paul \textit{et al}.~\cite{paul2022vision}  further integrates and expands the scope of robustness analysis to elucidate the underlying factors behind Vits' superior robustness.
Raghu \textit{et al}.~\cite{raghu2021vision}  conducted a detailed analysis of the differences between ViT and CNNs from various perspectives, providing insightful understanding of their robustness disparities. Through comparisons with canonical CNNs, these studies demonstrate ViTs are generally more robust while also inspiring further work to improve the robustness of ViTs~\cite{vits_robust_4, vits_robust_5, vits_robust_6}. Building upon and different from previous robustness analysis works for ViTs, this work focuses on large kernel networks~\cite{convmixer, convnext, convnext_v2, replk, slak}, which have demonstrated competitive performance with ViTs on mainstream tasks and attracted considerable interest recently. We design systematical experiments to evaluate their robustness and provide insights to analyze the underlying factors.

\section{Are Large Kernel ConvNets Robust ?}\label{sec3}

\begin{figure}[t]
\centering
%\vspace{-13mm} %调整纵向距离
\includegraphics[width=\linewidth]{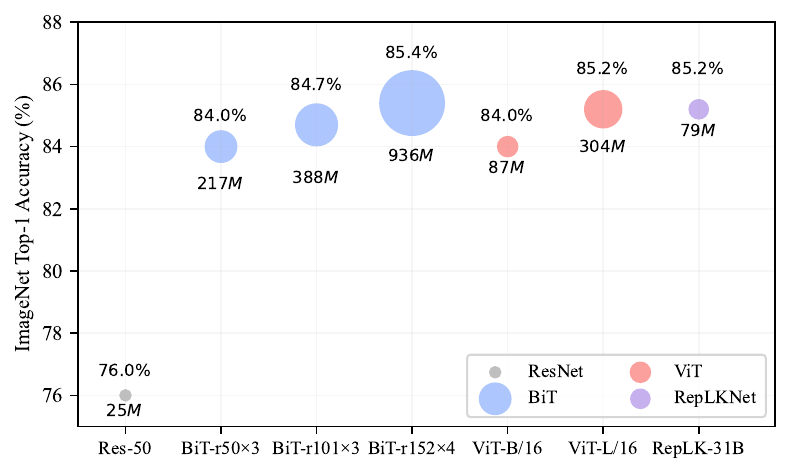} 
\vspace{-6.5mm} %调整纵向距离
\caption{\textbf{Model configurations.} We depict the model size and corresponding ImageNet-1k top-1 accuracy. We choose models with similar accuracy and parameter counts (except for resnet-50 as a baseline). All the reported variants were pre-trained on ImageNet-21k and then fine-tuned on ImageNet-1k.}
 \label{model}
\vspace{-3.9mm} %调整纵向距离
\end{figure}
In this section, we delve into the first two question: whether large kernel convnets are robust learners and how do they differ in terms of robustness from typical CNNs and ViTs. Specifically, we conduct an empirical study comparing large kernel convnets with their typical small kernel counterparts and ViT across six widely accepted robustness datasets. Section~\ref{sec3_1} provides a detailed introduction to the models' configurations, while section~\ref{sec3_2} presents and discusses the datasets and corresponding results.
\subsection{Model Configuration}\label{sec3_1}
We choose the representative work of large kernel convnets , RepLKNet~\cite{replk}, as the primary model for our experiments. RepLKNet is the first work to scale up kernel size to extremely large (i.e., $31\times31$) and has been applied to various vision tasks~\cite{slak, largekernel3d, link, lkteacher}. For comparison, we select ViT~\cite{vit} as a strong baseline. ViT has been proven to possess significant robustness by numerous studies~\cite{naseer2021intriguing,paul2022vision,shao2021adversarial,bhojanapalli2021understanding}, demonstrating a clear advantage over traditional convnets. For the comparison with typical convnets, we choose Big Transfer (BiT)~\cite{bit},  which has exceptional performance not only on ImageNet but also on various transfer learning scenarios~\cite{coco, vtab}. Since RepLKNet, BiT and ViT share similar pre-training strategies (such as using larger datasets like ImageNet-21K~\cite{imagenet}, extended pre-training schedules, and so on), they serve as excellent candidates for our comparison purposes. In addition, we add ResNet-50~\cite{resnet} as a basic baseline. We choose models
with similar accuracy and parameter counts (except for resnet-50 as a baseline). The parameter counts and top-1 accuracy on ImageNet-1K of different models are shown in Fig.~\ref{model}. Note that all the reported variants were initially pre-trained on ImageNet-21K and then fine-tuned on ImageNet-1K.

\subsection{Robustness Evaluation on Diverse Datasets}\label{sec3_2}
Next, we evaluate the performance of above model variants on six robustness benchmark datasets. These datasets assess the models' robustness from multiple dimensions including: \textbf{i)} natural adversarial; \textbf{ii)} common corruptions; \textbf{iii)} out-of-domain distribution; \textbf{iv)} common perturbations; \textbf{v)} semantic shifts; \textbf{vi)} background dependency. Their specific objectives and venues are summarized in Appendix~\ref{appendix_a}.
%Table~\ref{table-dataset}.

\begin{figure*}[t]
  \centering
\hspace{-38pt}
    \begin{subfigure}{0.339\textwidth}
      \centering   
      \includegraphics[width=1\linewidth]{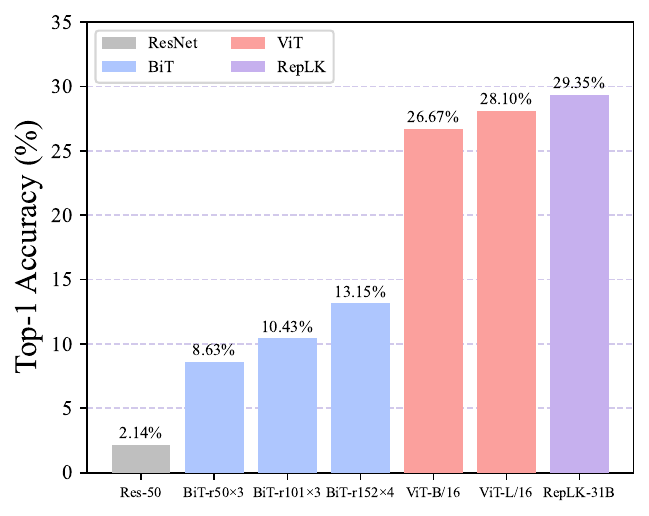}
        \caption{Accuracy comparison on ImageNet-A.}
        \label{fig_a}
    \end{subfigure}   %      \hfill  % 这个\hfill指令为插入弹性长度的空白，看情况选择加不加。
    %\hspace{-5pt}
    \begin{subfigure}{0.339\textwidth}
      \centering   
      \includegraphics[width=1\linewidth]{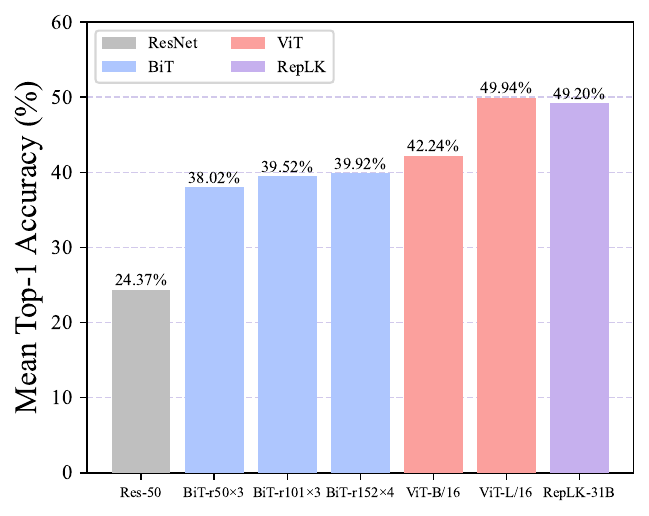}
        \caption{Accuracy comparison on ImageNet-C.}
        \label{fig_c}
    \end{subfigure}
    %\hspace{-5pt}
    \begin{subfigure}{0.339\textwidth}
      \centering   
      \includegraphics[width=1\linewidth]{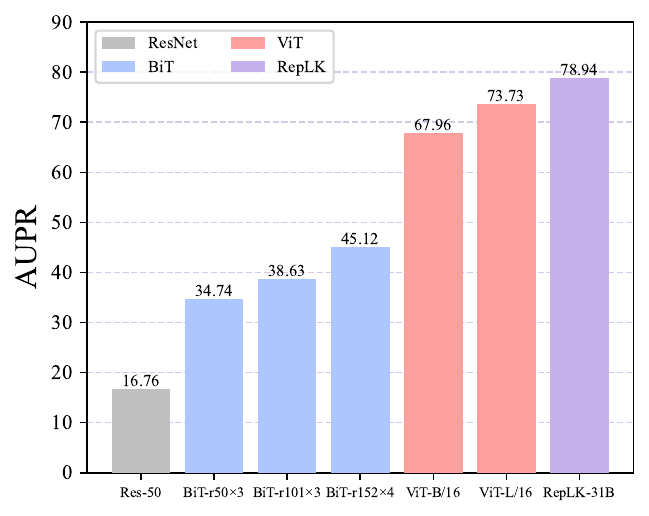}
        \caption{AUPR comparison on ImageNet-O.}
        \label{fig_o}
    \end{subfigure}
\hspace{-38pt}
    \vspace{-5pt}
\caption{\textbf{Comparison on ImageNet-A, ImageNet-C, and ImageNet-O.} For ImageNet-A, we report top-1 accuracy; For ImageNet-C, we report mean top-1 accuracy over all the 19 corruptions; For ImageNet-O, we report \textit{area under the precision-recall curve} (AUPR). Note that for all the metrics higher is better. RepLKNet-31B performs on par with or even better than ViT-L on ImageNet-A and ImageNet-O, demonstrating the strong robustness of large kernel convnets.
}
\label{fig_acor}
\end{figure*}

\textbf{ImageNet-A}~\cite{imagenet_ao} is a dataset of real-world adversarially filtered images that fool current ImageNet classifiers. In Fig.~\ref{fig_a}, we report the top-1 accuracy of ResNet, BiT, ViT and RepLKNet on the ImageNet-A dataset. In comparison to typical convnets, RepLKNet is shown to outperform ResNet and BiT by large margins. For instance, the top-1 accuracy of RepLKNet-31B is $2.8\times$ higher than BiT-m r101$\times$3. It even slightly surpasses ViT-L, which has a much larger model size, indicating that large kernel convnets possess strong robustness against natural adversarial challenges.

%\vspace{-10pt}

\textbf{ImageNet-C}~\cite{imagenet_cp} comprises 15 categories of algorithmically generated corruptions and an additional four general corruption types, resulting in a total of 19 corruption categories.
Each corruption type has five severity levels, ranging from negligible to pulverizing. We evaluate all 19 corruptions at their maximum severity level (5) and depict the mean top-1 accuracy in Fig.~\ref{fig_c}. We observed that RepLKNet performs on par with ViT-L and surpasses all typical convolutional networks and ViT-B/16. Considering the notably higher parameter count of ViT-L compared to RepLKNet, this further validates the robustness of large kernel convnets.

\begin{table*}[t]
\begin{floatrow}
\hspace{-8mm}
\capbtabbox{
\resizebox{0.21\textwidth}{!}{
 \begin{tabular}{l|c}
 \toprule
 Model / Method & mCE \\
 \midrule
 ResNet-50 & 76.7 \\
 BiT m-r101×3 & 58.3 \\
 DeepAugment + AugMix & 53.6 \\
 ViT-L/16 & 45.5 \\
 RepLKNet-31B & 36.5 \\
\bottomrule
 \end{tabular}}
}{
 \caption{mCEs (\%) of different models and methods on ImageNet-C (lower is better).}
 \label{tab:tb1}
}
\hspace{-6mm}

\capbtabbox{
\resizebox{0.21\textwidth}{!}{
 \begin{tabular}{l|c|c}
 \toprule
 Model / Method & mFR & mT5D\\
 \midrule
 ResNet-50 & 58.0 & 82.0\\
 BiT m-r101×3 & 50.0 & 76.7 \\
 AugMix & 37.4 & N/A \\
 ViT-L/16 & 33.1 & 50.2 \\
 RepLKNet-31B & 29.2 & 49.8 \\
\bottomrule
 \end{tabular}}
}{
 \caption{mFRs (\%) and mT5Ds (\%) comparison on ImageNet-P dataset (lower is better).}
 \label{table-p}
}
\hspace{2mm}

%\hspace{2mm}
\capbtabbox{
 \resizebox{0.472\textwidth}{!}{
 %\vspace{-15mm}
 \begin{tabular}{l|c|c|c|c}
 %\vspace{5mm}
 \toprule
 Model & Origin & Mixed-Same & Mixed-Rand & BG-Gap\\
 \midrule
 BiT m-r101×3 & 94.3 & 81.2 & 76.6 &4.6 \\
 ResNet-50 & 95.6 & 86.2 &78.9 & 7.3\\
 ViT-L/16 & 96.7 & 88.5 & 81.7 &6.8 \\
 RepLKNet-31B & 97.3 & 92.3 & 88.2 &4.1 \\
\bottomrule
 \end{tabular}}
}{
 \caption{Top-1 acc comparison on ImageNet-9. BG-Gap=Mixed-Same - Mixed-Rand. It measures how impactful background correlations are when correct-labeled foregrounds exist.}
 \label{tab-9}
}
\hspace{-16mm}
\end{floatrow}
\end{table*}

Moreover, \citep{imagenet_cp} propose mean corruption error (mCE) to quantify the robustness factors of a model on ImageNet-C. We follow the same evaluation approach and report mCE comparison in Table~\ref{tab:tb1}. We additionally add DeepAugment~\cite{imagenet_r} and AugMix~\cite{augmix}, which are specifically aimed at enhancing the model's robustness against corruptions observed in ImageNet-C. Surprisingly, RepLKNet outperforms other methods with a clear gap.
\begin{figure}[t]
\centering
%\vspace{-13mm} %调整纵向距离
\includegraphics[width=0.7\linewidth]{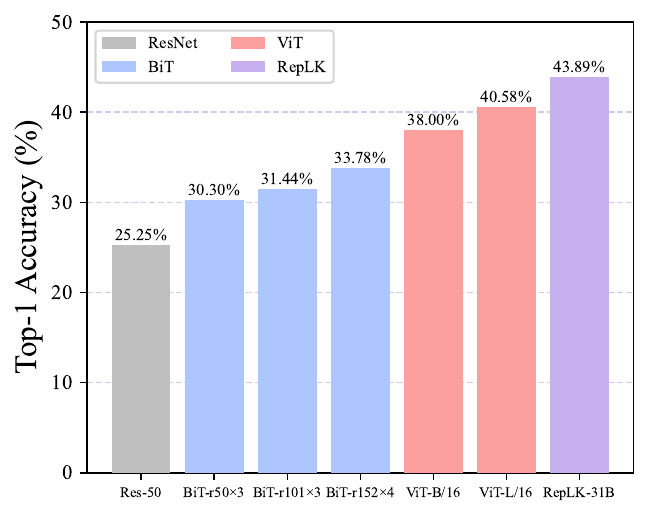} 
\vspace{-3.5mm} %调整纵向距离
\caption{\textbf{Accuracy comparison on ImageNet-R.} We report top-1 accuracy for all the model variants. RepLKNet-31B still outperforms ViT-B/L by large margins.}
 \label{fig_r}
%\vspace{-4.9mm} %调整纵向距离
\end{figure}%\vspace{-10pt}

\textbf{ImageNet-O}~\cite{imagenet_ao} is a dataset of adversarially filtered examples for ImageNet out-of-distribution detectors. Following \cite{imagenet_ao}, we use area under the precision-recall curve (AUPR) as the evaluation metric for ImageNet-O. As shown in Fig.~\ref{fig_o}, RepLKNet outperforms all the other model variants by clear margins, demonstrating large kernel convnets' superior robustness in anomaly detection.

\textbf{ImageNet-P}~\cite{imagenet_cp} consists of 10 types of common perturbations. ImageNet-P differs from ImageNet-C in that it generates perturbation sequences from each ImageNet validation image. The perturbations are subtly nuanced, affecting a smaller number of pixels within the images. We follow \citep{imagenet_cp} to use mean flip rate (mFR) and mean top-5 distance (mT5D) as the standard metrics to evaluate models' robustness. For brevity, we omit the detailed formulation of mFR and mT5D here. As shown in Table~\ref{table-p}, the robustness of RepLKNet has once again been confirmed to be better than BiT, ViT and AugMix as well.

\textbf{ImageNet-R}~\cite{imagenet_r} contains various artistic renditions of 200 classes from the ImageNet-1K dataset. We report top-1 accuracy comparison on ImageNet-R in Fig.~\ref{fig_r}, exhibiting that RepLKNet’s robustness to domain adaptation is better than that of BiT and ViT.

\textbf{ImageNet-9}~\cite{imagenet_9} helps disentangle the impacts of foreground and background signals on classification. It measures the model's robustness towards background changes. As shown in Table~\ref{tab-9}, we report four metrics to assess the model's robustness to background changes. \textit{Origin} refers to the original accuracy without modifying background. \textit{Mixed-Same} involves replacing the original background with a random one from \textbf{the same} class. \textit{Mixed-Rand} means replacing the original background with a random background from \textbf{a random} class. \textit{BG-Gap = } \textit{Mixed-Same} - \textit{Mixed-Rand}, which measures how impactful background correlations are when correct-labeled foregrounds exist. The results reveals that RepLKNet surpasses BiT and ViT consistently on the first three metrics. Additionally, the BG-Gap for RepLKNet is much smaller, this suggests that large kernel convnets are less sensitive to background modifications, exhibiting stronger background robustness.

\begin{figure*}[t]
  \centering
  \hspace{-28pt}
    \begin{subfigure}{0.345\textwidth}
      \centering   
      \includegraphics[width=1\linewidth]{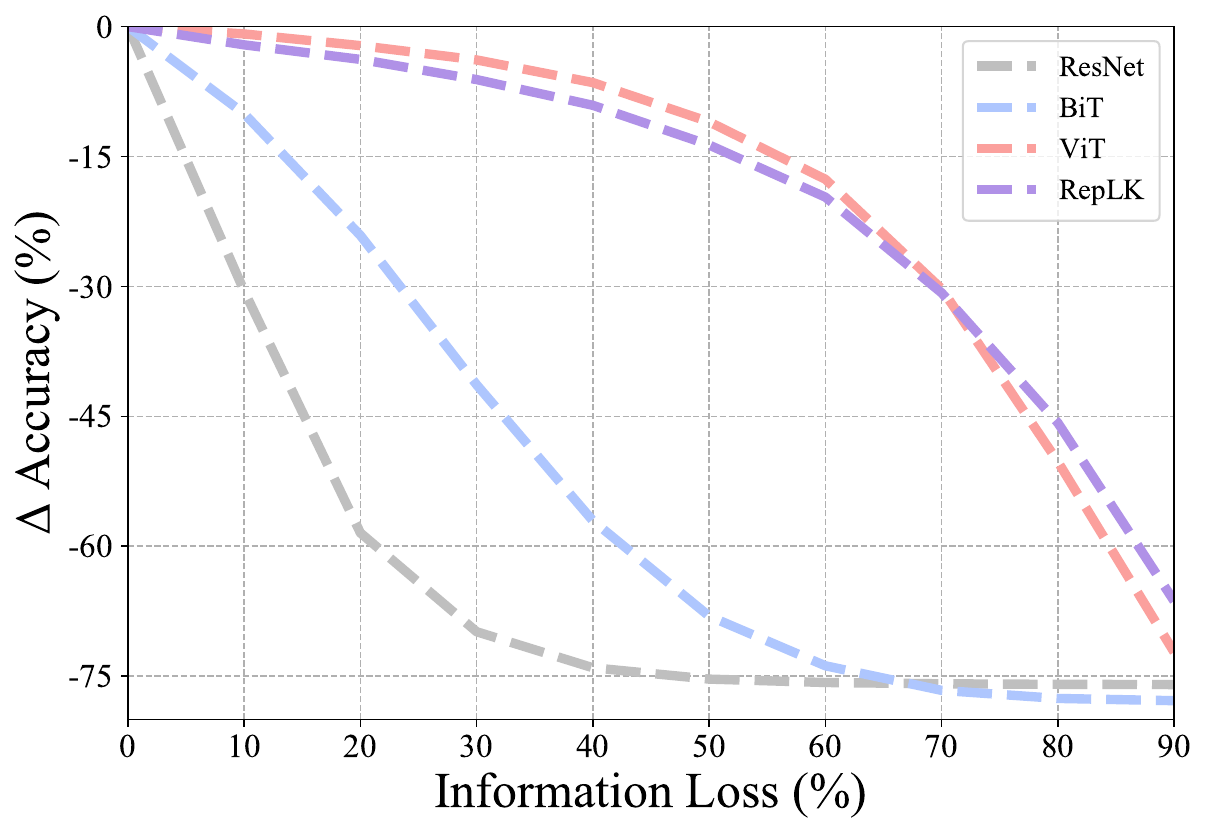}
        \caption{Random patch-drop.}
        \label{fig_random}
    \end{subfigure}   %      \hfill  % 这个\hfill指令为插入弹性长度的空白，看情况选择加不加。
    %\hspace{39pt}
    \hspace{-5pt}
    \begin{subfigure}{0.33\textwidth}
      \centering   
      \includegraphics[width=1\linewidth]{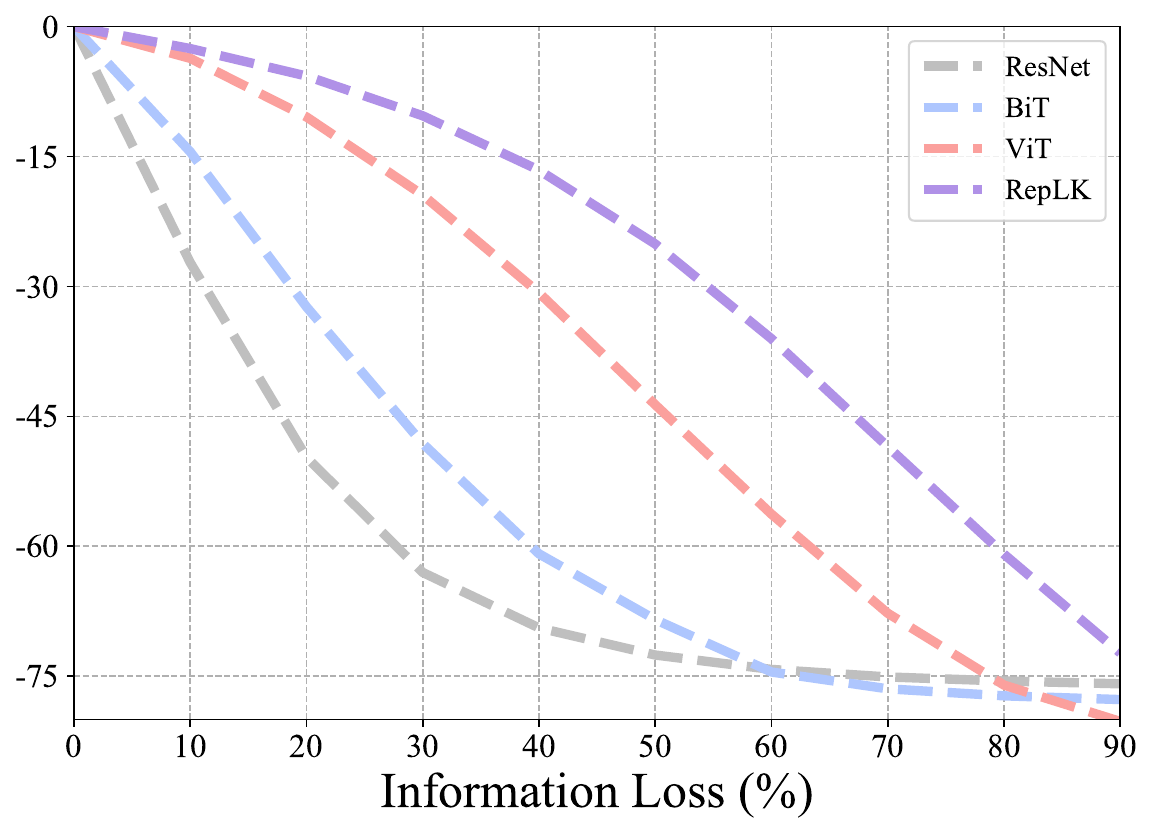}
        \caption{Salient patch-drop.}
        \label{fig_salient}
    \end{subfigure}
    \hspace{-5pt}
    \begin{subfigure}{0.33\textwidth}
      \centering   
      \includegraphics[width=1\linewidth]{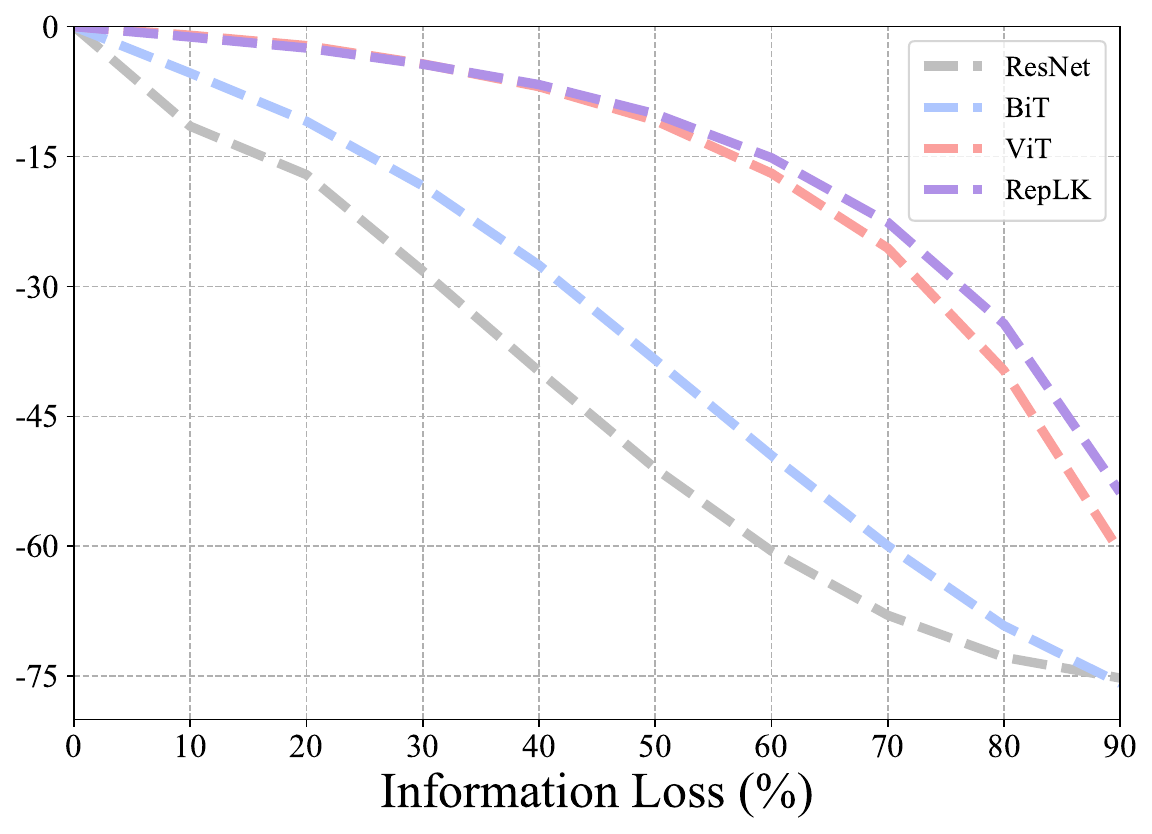}
        \caption{Non-salient path-drop.}
        \label{fig_nonsalient}
    \end{subfigure}
    \hspace{-28pt}
\caption{
\label{fig_occlusion}
\textbf{Occlusion robustness comparison under different settings.} We use random drop, salient drop and non-salient drop to evaluate corresponding occlusion robustness. We report top-1 accuracy drop under different information loss ratios, ranging from 10\% to 90\%. RepLKNet are more robust to extreme occlusion scenarios and more importantly, they outperform ViT for salient occlusion remarkably. 
}
\vspace{-4.9mm} %调整纵向距离
\end{figure*}

\section{Why are Large Kernel ConvNets Robust?}\label{sec4}
In this section, we thoroughly investigate the robustness of large kernel convnet from both quantitative and qualitative perspectives. To be specific, we systematically design nine distinctive experiments to conduct a comprehensive and in-depth analysis of its key properties like occlusion-invariance and kernel attention patterns, providing multi-dimensional insights into its strong robustness. 

\subsection{Occlusion Invariance}

\textit{Occulusion Invariance} matters in science and deep learning~\cite{occlusion1,occlusion2}. It plays a crucial role in human vision's robustness: for instance, we can imagine the occluded parts based on texture or shape even when most of the object is invisible, maintaining consistent judgement under high occlusion ratios. To this end, we delve into the occlusion invariance of large kernel networks under various scenarios, where some or most of the image content is missing.
\begin{figure}[t]
\centering
%\vspace{-13mm} %调整纵向距离
\includegraphics[width=\linewidth]{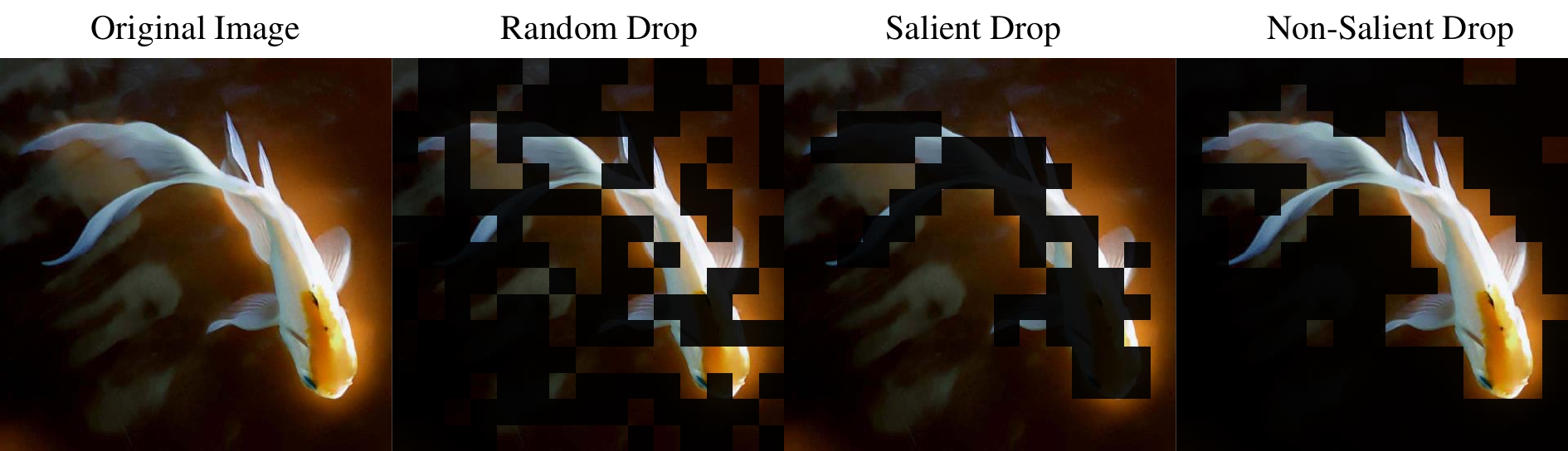} 
%\vspace{-6.5mm} %调整纵向距离
\caption{\textbf{Illustration of patch drop.} We depict an example image of different occlusion types: random, salient and non-salient. The pixel values in the occluded (black) areas are assigned to be zero.}
 \label{fig_example}
\vspace{-4.9mm} %调整纵向距离
\end{figure}

\textbf{Occlusion modeling.} Formally, we define occlusion in a patch-manner: we firstly divide the input image ${\rm X} \in \mathbb{R}^{\rm H \times W \times C}$ into non-overlapping flattened patches $\rvx = [\rvx_1, \rvx_2, ..., \rvx_{\rm N}]$, where $\rvx_i \in \mathbb{R}^{\rm P^2 C}$ according to the patch size $\rm P$, and $\rm N = (H\times W)/P^2$ is the number of patches. We choose a subset of the total image patches, $\rm M<N$, and set pixel values of these patches to zero to generate an occluded image, denoted as $\rvx'$. We feed $\rvx'$ into different models and measure their occlusion invariance by the top-1 accuracy on $\rvx'$. We conduct experiments with three occlusion types: i) random drop; ii) salient (foreground) drop; 3) non-salient (background) drop. An illustration is provided in Fig.~\ref{fig_example}.

\textbf{Random drop.} Following ViT~\cite{vit}, we partition images into $16\times16$ patches, then for a typical image with resolution $224\times224$, there will be 196 patches. This design makes a convenient and fair comparison with ViT without altering its input format. We randomly mask different proportions of the image, ranging from 10\% to 90\%. As shown in Fig.~\ref{fig_random}, typical convnets suffer catastrophic degradation when the occlusion ratio reaches 50\%, whereas RepLKNet and ViT exhibit clear advantages.  Moreover, as the occlusion ratio becomes extremely high, RepLKNet gradually outperforms ViT, indicating that large kernel convnets are more robust to extreme occlusion scenarios.

\begin{figure*}[t]
  \centering
  \hspace{-28pt}
    \begin{subfigure}{0.31\textwidth}
      \centering   
      \includegraphics[width=1\linewidth]{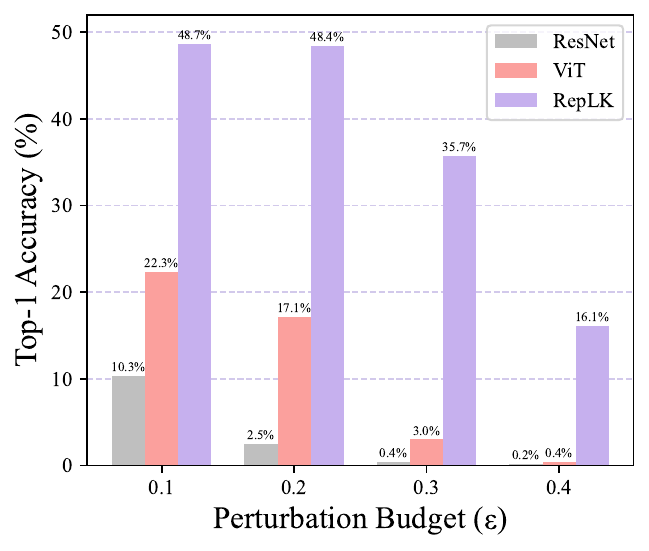}
        \caption{FGSM Attack.}
        \label{fig_fgsm}
    \end{subfigure}   %      \hfill  % 这个\hfill指令为插入弹性长度的空白，看情况选择加不加。
    \hspace{-5pt}
    \begin{subfigure}{0.31\textwidth}
      \centering   
      \includegraphics[width=1\linewidth]{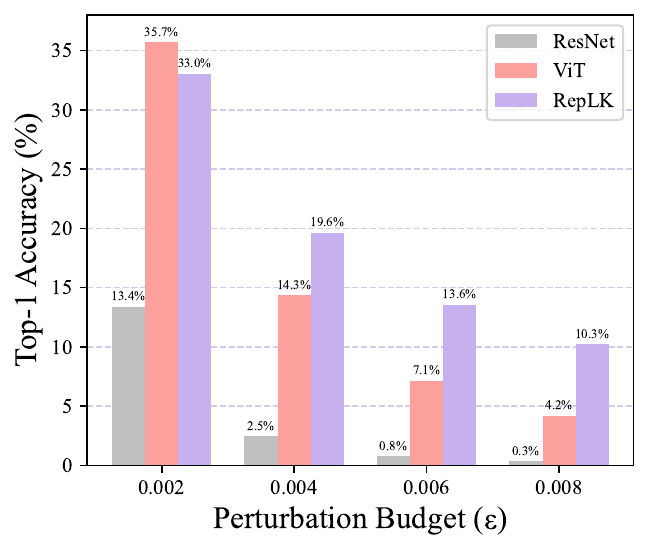}
        \caption{PGD Attack (step=5).}
        \label{fig_pgd}
    \end{subfigure}
    \hspace{-5pt}
    \begin{subfigure}{0.38\textwidth}
      \centering 
      %\hspace{-10pt}
      \includegraphics[width=1\linewidth]{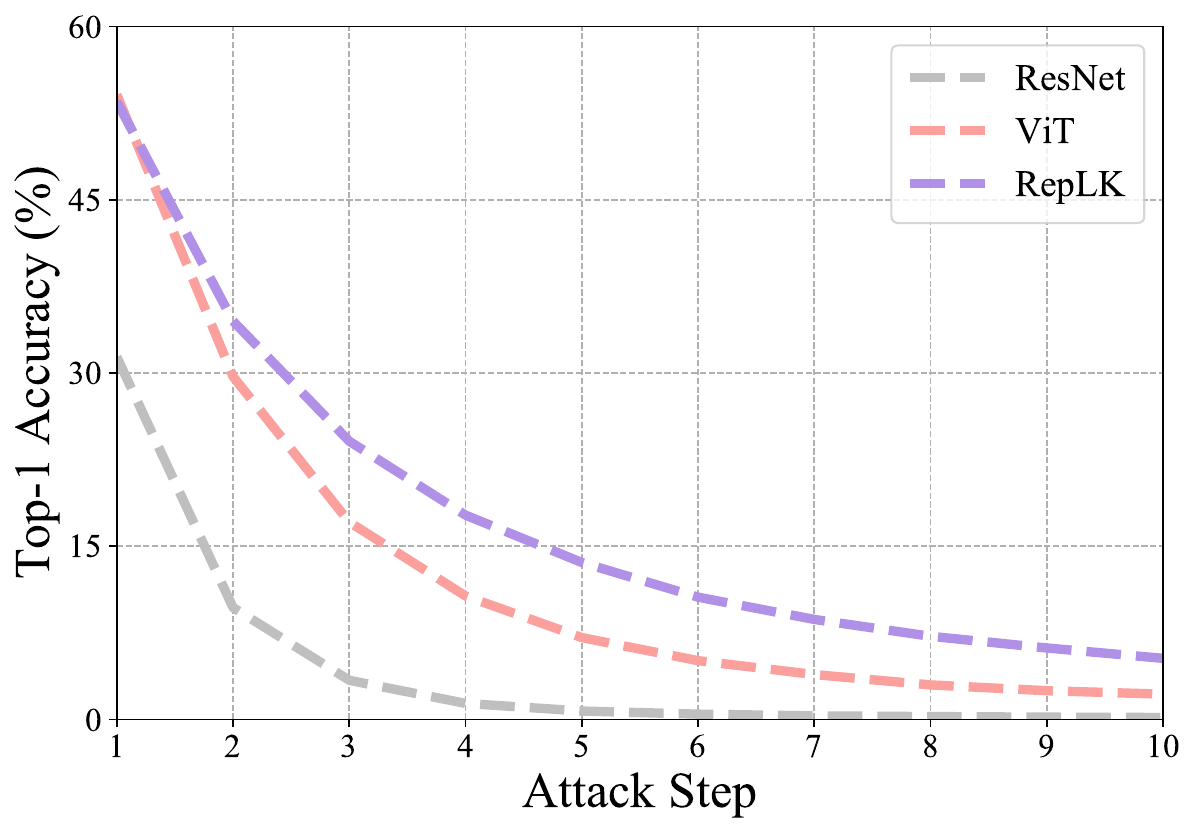}
        \caption{PGD Attack ($\varepsilon=0.006$).}
        \label{fig_pgd_step}
    \end{subfigure}
    \hspace{-28pt}
\caption{
\label{fig_adversarial}
\textbf{Comparison under different adversarial attacks.} (a) For FGSM attack, we apply with increasing perturbation budget $\varepsilon$. (b) For PGD attack, we fix the attack step to 5 while varying the perturbation budget. (c) For PGD attack, we fix $\varepsilon$ to 0.006, then gradually increase the attack step. RepLKNet consistently exceeds both ViT and ResNet. 
}
\vspace{-4.9mm} %调整纵向距离
\end{figure*}

\textbf{Salient (foreground) drop.} Foreground objects play a decisive role in visual recognition, hence investigating robustness against foreground occlusion is of vital importance. Firstly, we use a pretrained DINO~\cite{dino} to detect objects and identify salient regions. Then we mask a subset of patches encompassing the top K\% of foreground information. It's important to note that this K\% doesn't always equate to the pixel percentage. For instance, an image's 70\% foreground data might be contained in just 20\% pixels. As shown in Fig.~\ref{fig_salient}, RepLKNet exhibits remarkable robustness to salient occlusion, even surpassing ViT noticeably. We reckon this is the key reason for its substantial advantage over ViT on background-dependency dataset (i.e., ImageNet-9).

\textbf{Non-salient (background) drop.} Contrary to the above salient setting, we further select patches containing the lowest K\% of foreground information and mask them. Fig.~\ref{fig_nonsalient} shows that RepLKNet also has superior robustness to non-salient occlusion.

\subsection{Robustness to Adversarial Attack}
Previous studies reveal that even minor adversarial perturbations can substantially alter the decision boundary of neural networks~\cite{hold_me_tight}  and investigating adversarial attacks plays a crucial role in enhancing the robustness of models~\cite{madry2017towards}. Hence in this subsection, we step further into the performance of large kernel convnets against adversarial attacks. 

We use both single-step and multi-step sample specific attacks, Fast Gradient Sign Method (FGSM)~\cite{fgsm} and Projected Gradient Attack (PGD)~\cite{pgd} respectively. Note that we only compare RepLKNet with ResNet and ViT for this experiment and omit BiT here as it continues to exhibit performance between ResNet and ViT, similar to previous sections' observations.

We conduct this experiment with three variants. \textbf{i)} For FGSM attack, we apply with an $l_{\infty}$ perturbation of increasing budget $\varepsilon$, ranging from 0.1 to 0.4. As shown in Fig.~\ref{fig_fgsm}, ViT and ResNet becomes comletely destroyed when the budget grows to 0.3, while RepLKNet still holds a 35.7\% top-1 accuracy; \textbf{ii)} For PGD attack, we first fix the attack step to be 5, varing the perturbation budget $\varepsilon$ from 0.002 to 0.008. Results in Fig.~\ref{fig_pgd} exhibit that when the adversarial attacks are strong, RepLKNet outperforms ViT by clear margins; \textbf{iii)} Then for PGD attack, we fix the perturbation budget $\varepsilon$ to be 0.006, and we gradually increase the step to observe the robustness trend of different models in Fig.~\ref{fig_pgd_step}. Similarly, RepLKNet consistently exceeds both ViT and ResNet. Note that for all the experiments, we normalize pixel values using default ImageNet mean and standard deviation. The above experiments across three settings point to the same conclusion: large kernel network possesses broader and stronger adversarial robustness, and this could potentially explain why RepLKNet performs better on natural adversarial dataset (i.e., ImageNet-A).
\begin{table}[t]
	\begin{center}
   \caption{\textbf{Comparison under TAIG attack.} We increase the perturbation budget $\epsilon$ from 0.03 to 0.1. The attack success rate is reported, note that lower is better.} 
  \label{table_TAIG}
		\small
		\begin{tabular}{l|c|c|c}
			\toprule
			Model			&	$\epsilon=0.03$ &	$\epsilon=0.06$	&	$\epsilon=0.1$ \\
                \midrule

                ResNet-50      &   47.7    &   65.2       & 67.9 \\
                
              BiT-r152$\times$4      &   44.6    &   60.2        & 63.1 \\
               ViT-B     &   62.3    &  73.3       & 75.6 \\
ViT-L      &   54.9    &   67.9        & 69.8 \\
  \rowcolor[rgb]{0.95,0.95,0.95}	RepLKNet-31B    &  \textbf{39.2}  &   \textbf{56.1}      & \textbf{59.3} \\
			\bottomrule
		\end{tabular}	
  \end{center}
\vspace{-0.25in}
\end{table}

In addition to the classic adversarial attacks FGSM and PGD, we further evaluate with a modern strong adversarial attack TAIG~\cite{taig}. Specifically, for convnets (i.e., ResNet, BiT, RepLKNet), ViT-B is selected as the surrogate model to generate adversarial examples, for ViTs, RepLKNet-31B is selected as the surrogate model. We increase the $\epsilon$ from 0.03 to 0.1 to gradually increase the attack budget. We report the attack success rate (lower is better). As shown in Table~\ref{table_TAIG}, large kernel network still behaves better than typical small kernel convnets and ViTs, showing its inherent robustness against adversarial attacks.

\begin{figure}[t]
\centering
%\vspace{-13mm} %调整纵向距离
\includegraphics[width=0.933\linewidth]{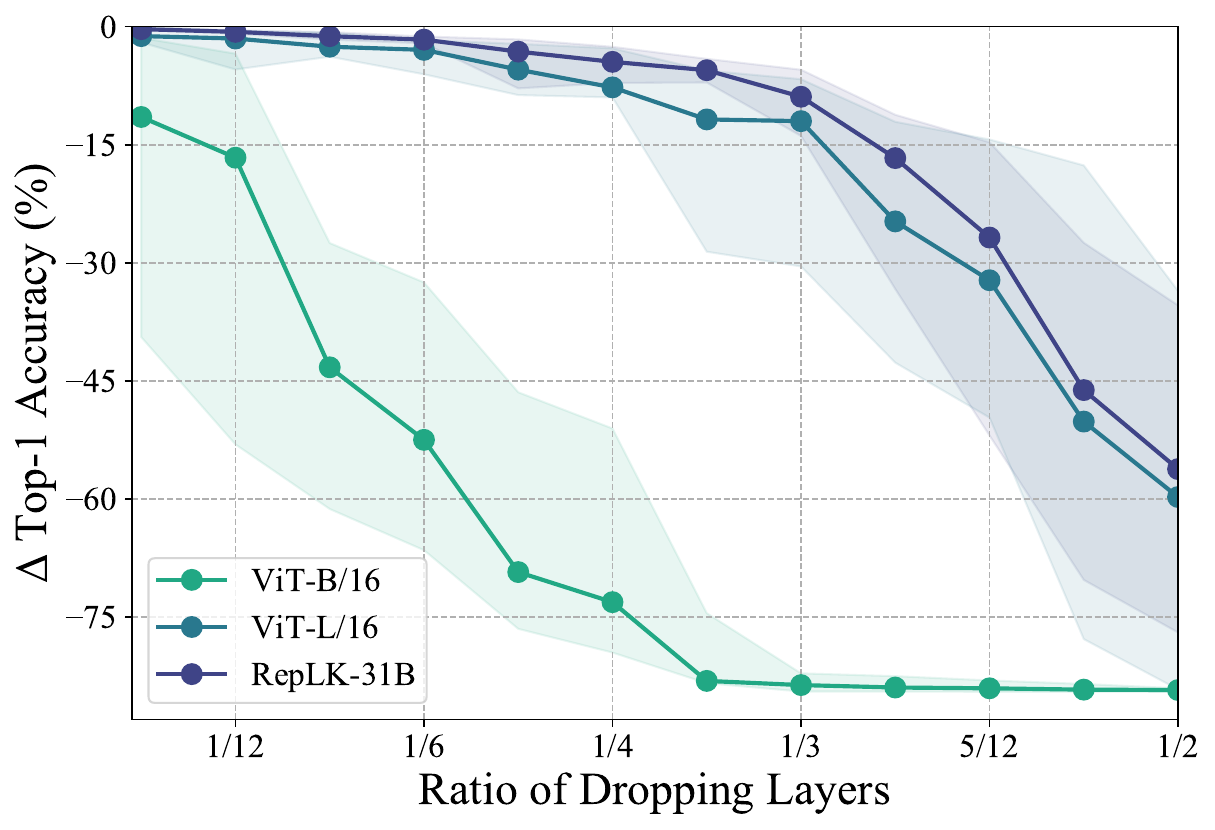} 
%\vspace{-6.5mm} %调整纵向距离
\caption{\textbf{Lesion study.} Evaluation of RepLKNet and ViT-B/L when $n(n\geqq1)$ blocks are removed from the model after training. For each n, results are from 10 independent samples of $n$ blocks and we show the average accuracy (line) and min/max (shaded area) across samples. %We observe that RepLKNet consistently outperforms ViTs against model perturbations.
}
 \label{fig_lesion}
\vspace{-4.9mm} %调整纵向距离
\end{figure}

\subsection{Robustness to Model Perturbations}
Previous studies~\cite{greff2016highway,veit2016residual,bhojanapalli2021understanding} have shown that layers in residual networks exhibit a large amount of redundancy, and that almost any individual layer can be removed after training without hurting performance. On the other hand, it is observed that identity shortcut is of vital importance especially for networks with very large kernels~\cite{replk}. To comprehend the information flow in large kernel convnets, we conduct a lesion study where we remove several blocks from an already trained network during inference, such that information has to flow through the skip connection.

Specifically, we randomly remove $n (n\geqq1)$ blocks and report the changes in top-1 accuracy. For each $n$, results are from 10 independent samples of $n$ blocks and we show the average accuracy (line) and min/max (shaded area) across samples. Given that the number of blocks in ViT-B is only half of that in RepLKNet and ViT-L, we use the proportion of blocks dropped as the x-axis in Fig.~\ref{fig_lesion}. We omit ResNet here as it is much more fragile than ViT-B.

We observe that ViT-B is highly sensitive to layer removal, with the model's accuracy nearing zero after removing just $\frac{1}{3}$ of the blocks. In contrast, RepLKNet and ViT-L can still maintain an accuracy of around $75\%$ at that dropping ratio. Moreover, we notice that despite having a smaller capacity, RepLKNet consistently outperforms ViT-L in terms of accuracy under the same block dropping ratios. This suggests that large kernel networks are more robust to model perturbations, which may explain their robustness against common perturbations (i.e., ImageNet-P).
\begin{figure}[t]
\centering
%\vspace{-13mm} %调整纵向距离
\includegraphics[width=0.91\linewidth]{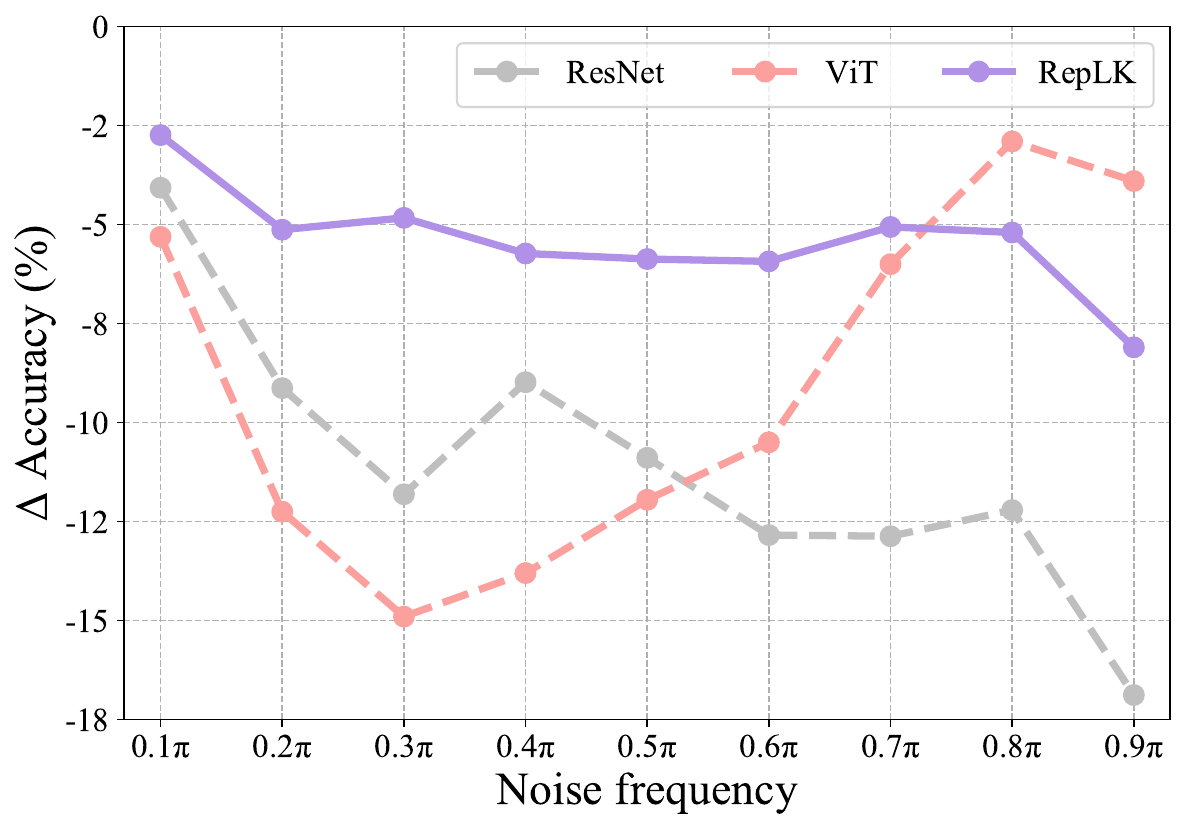} 
%\vspace{-6.5mm} %调整纵向距离
\caption{\textbf{Robustness to noise frequency.} We measure the decrease in accuracy against frequency-based random noise. ResNet is susceptible to high-frequency noise, while ViT is more affected by low-frequency noise. In a sharp contrast, RepLKNet maintains consistent robustness against noise across all frequencies.}
 \label{fig_noise_frequency}
\vspace{-4.9mm} %调整纵向距离
\end{figure}
\begin{table*}[h]
%\vspace{-5.5mm}
	\begin{center}
  \caption{\textbf{Ablation on kernel size.} We replace all large kernel convolutions in RepLKNet-31B with 3x3 small kernels.   The metrics are the same as Section~\ref{sec3}\&\ref{sec4}.  Reducing the kernel size from 31 to 3 significantly degrades the robustness of RepLKNet across various metrics.}
   \label{table-kernel-ablation}
		\small
		\scalebox{1.0}{\begin{tabular}{l|c|c|c|c|c|c}
 \toprule
 Model & ImageNet-A & ImageNet-C & ImageNet-R & ImageNet-O & Salient-Drop-50\% & Noise-$0.7\pi$\\
 \midrule
 RepLKNet-31B & 29.4 \textcolor[rgb]{0.8,0.1,0.1}{(+2.7)}  & 49.2 \textcolor[rgb]{0.8,0.1,0.1}{(+7.0)} & 43.9 \textcolor[rgb]{0.8,0.1,0.1}{(+5.9)}&78.9 \textcolor[rgb]{0.8,0.1,0.1}{(+10.9)}&-25.1 \textcolor[rgb]{0.8,0.1,0.1}{(+19.3)}&-5.0 \textcolor[rgb]{0.8,0.1,0.1}{(+1.6)}\\
 ViT-B &  26.7 \textcolor[rgb]{0.7,0.7,0.7}{(-- --)}&42.2 \textcolor[rgb]{0.7,0.7,0.7}{(-- --)}&38.0 \textcolor[rgb]{0.7,0.7,0.7}{(-- --)}&68.0 \textcolor[rgb]{0.7,0.7,0.7}{(-- --)}&-44.4 \textcolor[rgb]{0.7,0.7,0.7}{(-- --)}&-6.6 \textcolor[rgb]{0.7,0.7,0.7}{(-- --)}\\
 RepLKNet-3B & 14.6 \textcolor[rgb]{0.1,0.6,0.1}{(-12.1)}  & 39.1 \textcolor[rgb]{0.1,0.6,0.1}{(-3.1)} &32.9 \textcolor[rgb]{0.1,0.6,0.1}{(-5.1)}&51.6 \textcolor[rgb]{0.1,0.6,0.1}{(-16.4)}&-61.8 \textcolor[rgb]{0.1,0.6,0.1}{(-17.4)}&-10.6 \textcolor[rgb]{0.1,0.6,0.1}{(-4.0)}\\
\bottomrule
 \end{tabular}}
	\end{center}
%\vspace{-3.5mm}
\end{table*}

\subsection{Robustness to Noise Frequency}

To deepen the understanding of large kernel convnet's strong robustness, we further analyze its robustness in frequency-domain. In particular, we subject the model to random noise attacks at varying frequencies and evaluate the accuracy drop. We normalize the frequency to be between $0.0\pi$ (center) and $1.0\pi$ (boundary). We use a frequency window size of $0.1\pi$ for frequency-based noise.

As shown in Fig.~\ref{fig_noise_frequency}, ResNet is highly susceptible to high-frequency noise, while ViT exhibits poorer performance against low-frequency noise. This phenomenon aligns well with previous study~\cite{how}. In contrast, RepLKNet consistently demonstrates robustness against noise across all frequency bands. For instance, for noise across $0.1\pi$ to $0.8\pi$ frequency range, RepLKNet consistently maintains the accuracy loss within 6\%, consistently outperforming ResNet and ViT.

\subsection{Large Kernel Size is the Key}

Since RepLKNet, BiT, and ViT share similar pre-training strategies, the primary distinction between them lies in the use of large kernel convolutions. Therefore, we proceed to explore the impact of kernel size on robustness. Specifically, we replace all large kernel convolutions in RepLKNet-31B with 3x3 small kernel convolutions (the same as kernel size used in BiT and ResNet), while maintaining the same data augmentation and training schedule. We train the modified model, RepLKNet-3B, and evaluate its robustness across multiple above-mentioned tasks. As shown in Table~\ref{table-kernel-ablation}, reducing the kernel size from 31 to 3 significantly degrades the robustness of RepLKNet across various metrics, leading to inferior performance compared to ViT-B. This finding highlights the critical role of large kernel convolution in enhancing model robustness.

\begin{figure}[t]
\centering
%\vspace{-13mm} %调整纵向距离
\includegraphics[width=0.91\linewidth]{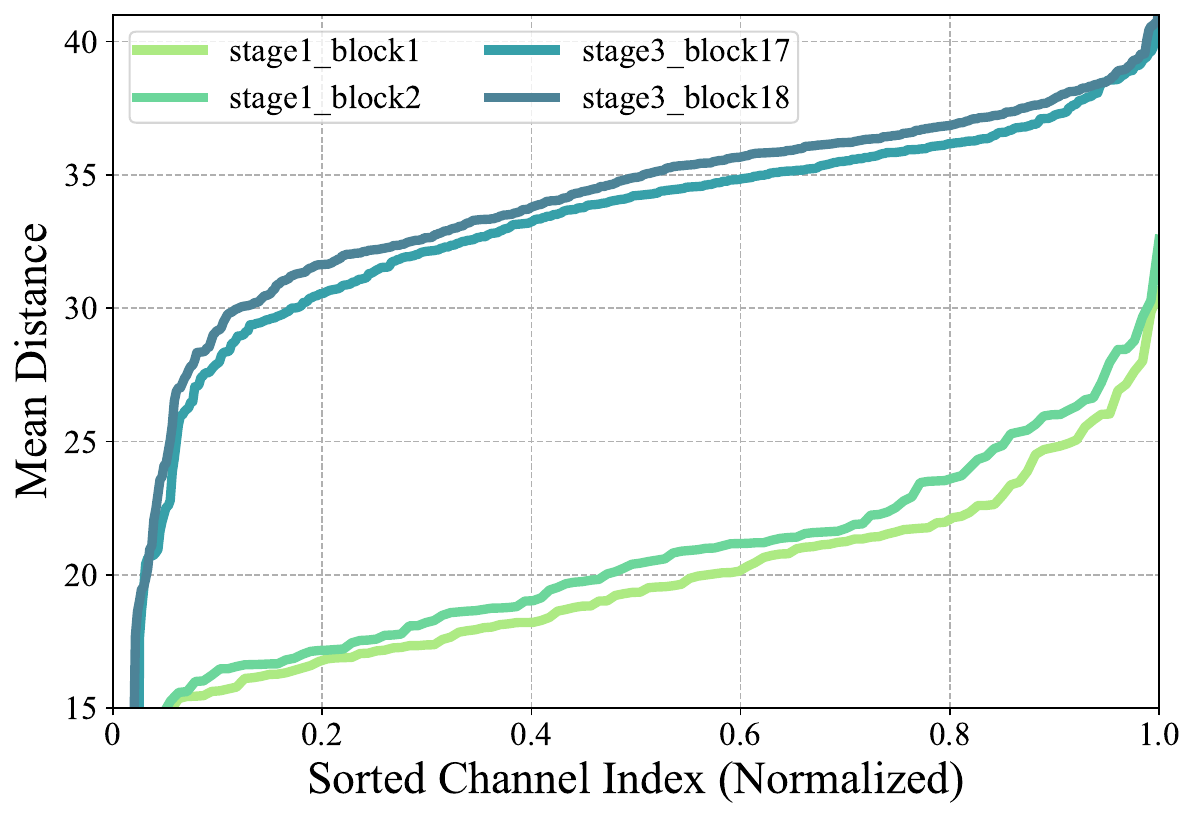} 
%\vspace{-6.5mm} %调整纵向距离
\caption{\textbf{Kernel attention distance of different layers.} We calculate RepLKNet's sorted kernel attention distance in the first two layers of stage-1 and the last two layers of stage-3. It also tends to aggregate both local and global information at shallow layers, while focusing more on global information at deeper layers.}
 \label{fig_kernel_atten_dis}
\vspace{-4.9mm} %调整纵向距离
\end{figure}

\subsection{Local and Global Kernel Attention}
Then we dive into the kernel attention pattern to explore the property of large kernels. Facilitated by the self-attention mechanism, ViT can aggregate global information through the global interactions between all tokens. However, global attention is not always optimal. Raghu \textit{et al.}~\cite{raghu2021vision} revealed that powerful ViTs consider both local and global information at shallow layers, mainly focusing on global information at deeper layers. This is in contrast to typical CNNs~\cite{vgg, resnet}, which are hardcoded to attend only locally in all layers. Since large kernel convnets have large receptive field comparable to ViTs~\cite{replk,slak}, we conjecture that they may exhibit similar attention patterns with ViT, attending to both local and global information at shallow layers while mainly global in deep layers.

To verify, we calculate the average kernel attention distance of different layers in RepLKNet. Specifically, for each kernel, we measure the Euclidean distance from every kernel position to the center of the kernel,  we weight this distance by the absolute value of the corresponding parameter, and then we normalize the sum of all parameters to get this kernel's attention distance. Since each kernel has multiple channels, we sort them by the kernel attention distance.

We depict the kernel attention distance of the first two layers of stage1 and the last two layers of stage3 in Fig.~\ref{fig_kernel_atten_dis}, since they have different channel dims, we normalize the channel index for a better comparison. Interestingly, RepLKNet also tends to aggregate both local and global information at shallow layers, while focusing more on global information at deeper layers. This essentially suggests that simultaneously aggregating local and global information can more effectively capture different levels of information in images, thereby resulting in more powerful and robust performance.

\begin{figure}[t]
\centering
%\vspace{-13mm} %调整纵向距离
\includegraphics[width=0.9\linewidth]{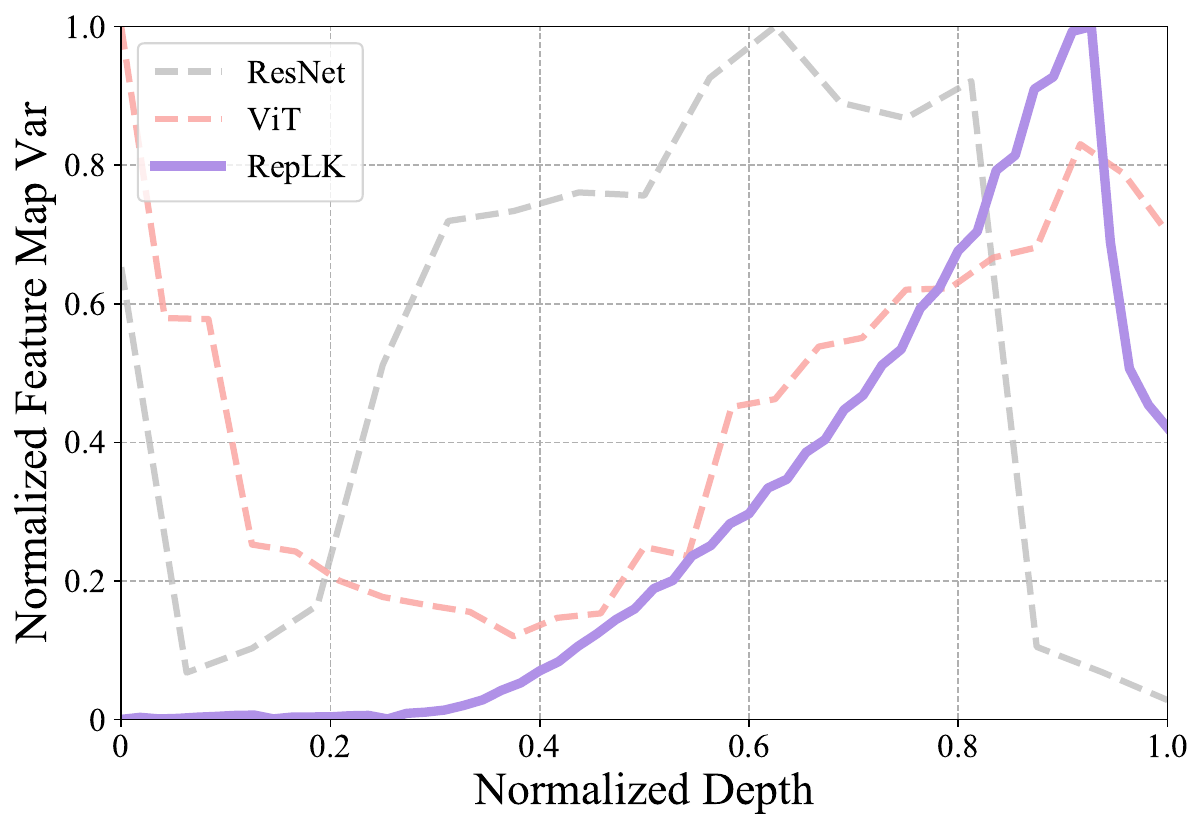} 
%\vspace{-6.5mm} %调整纵向距离
\caption{\textbf{Feature map variance change curve.} We measure the normalized feature map variance given the same input. Compared with ResNet and ViT, RepLKNet has much lower feature map variance in the very early layers and the overall curve changes in a simpler and more stable manner.}
 \label{fig_featuremap_var}
\vspace{-4.9mm} %调整纵向距离
\end{figure}

\subsection{Stable Feature Map Variance}\label{sec_featuremap}
Further, we investigate the stability of the feature maps of large kernel networks. Specifically, we feed a batch of ImageNet validation images into different networks, the batch size is set to 64 and we use regular validation augmentation (only center crop and normalization). Then we calculate the normalized feature map variance layer by layer.
\begin{table*}[h]
	\begin{center}
   \caption{\textbf{More results of large kernel convnets.} We add ConvNeXt as another large kernel convnet, and scale up the model size to large to demonstrate the scaling properties. *Note that ImageNet-C's image size is only $224\times224$.} 
  \label{table_convnext}
		\scalebox{0.91}{
		\begin{tabular}{l|c|c|c|c|c}
			\toprule
			Model		&Kernel Size	&	ImageNet &	ImageNet-A	&	ImageNet-R &ImageNet-C* \\
                \midrule
              BiT-r152$\times$4    &$3\times3$  &   85.4    &  13.2       & 33.8&39.9 \\
               ViT-B     & N/A&  84.0    &  26.7       & 38.0&42.2 \\
ViT-L      &  N/A& 85.2    &   28.1      & 40.6&49.9 \\
 
   ConvNeXt-B   &$7\times7$ &  85.8  &  33.9  &45.5&53.2 \\
   ConvNeXt-L   &$7\times7$ &  86.6  &  38.7  &47.6&56.5 \\
    RepLKNet-31B   &$31\times31$ &  85.2  &  29.4  &43.9&49.2 \\
   RepLKNet-31L   &$31\times31$ &86.6&39.6&49.1&52.6 \\
			\bottomrule
		\end{tabular}	}
  \end{center}
%\vspace{-0.25in}
\end{table*}

As shown in Fig.~\ref{fig_featuremap_var}, RepLKNet differs from the other two networks in two distinct aspects: \textbf{i)} It is very stable in the early stages. For instance, when the normalized depth is smaller than 0.4, the variance of RepLKNet's feature maps remains at a very low level, while ViT and ResNet tend to have a large variance from the very beginning; \textbf{ii)} The variance change in a simple and coherent manner. RepLKNet rises and falls slowly, changing in a simple and stable way throughout the process, while ResNet and ViT experience sharp fluctuations.

\subsection{More Large Kernel ConvNets' Results}
In order to further verify the effect of large kernel convolution on robustness and its scaling properties, we add another modern strong large kernel convnet, ConvNeXt~\cite{convnext}, and increase the model size of RepLKNet and ConvNeXt from base to large for comparison. As shown in Table~\ref{table_convnext}, ConvNeXt also demonstrates strong robustness, and its robustness is further improved when the model size increases. As for RepLKNet-31L, we use the $384\times384$ pretrained model from the official release, and it achieves consistent improvements on ImageNet, ImageNet-A and ImageNet-R. For ImageNet-C, this dataset's images are all $224\times224$, evaluating a $384\times384$ model on this dataset will cause inconsistent input distribution, thus the improvement on this dataset is less significant.

\subsection{How Large Can Make Strong Robustness?}
While ConvNeXt and RepLKNet have different large kernel sizes (i.e., $7\times7$ v.s. $31\times31$), they both exhibit strong robustness. It is natural to ask: \textit{how large kernel size can make strong robustness?} To answer this question, we conduct ablation studies by gradually increasing kernel sizes. 

Specifically, we train ConvNext-Tiny with different kernel sizes in a 120 epoch schedule on ImageNet-1K (for computation constraints, it is not affordable for us to conduct this ablation on ImageNet-21K). The only difference is the kernel size. As shown in Table~\ref{table_ablation_kernel}, there are three observations: \textbf{i)} scaling up kernels can bring consistent improvements both on ImageNet and robustness benchmarks; \textbf{ii)} basically, scaling up to $13\times13$ can make a favorable robustness, but continuing scaling up kernel size to $51\times51$ can bring further robustness ; \textbf{iii)} Although scaling up to extremely large can not bring significant improvements on ImageNet, it can bring better improvements on robustness.
\begin{table}[t]
	\begin{center}
   \caption{\textbf{Ablation of kernel sizes on robustness.} We gradually increase the kernel size of ConvNeXt-T from 3 to 51. Scaling up kernel size can make consistent improvements on robustness. Note that ImgN is short for ImageNet.} 
  \label{table_ablation_kernel}
		\scalebox{0.93}{
		\begin{tabular}{l|c|c|c|c}
			\toprule
			Model		&Kernel	&	ImgN &	ImgN-A	&	ImgN-R  \\
                \midrule
              ConvNeXt-3&$3\times3$&79.4&5.20&28.87\\
              ConvNeXt-7&$7\times7$&80.7&7.76&29.86\\
              ConvNeXt-13&$13\times13$&81.3&9.89&30.97\\
              ConvNeXt-31&$31\times31$&81.4&10.20&31.34\\
              ConvNeXt-51&$51\times51$&81.6&10.71&31.77\\
			\bottomrule
		\end{tabular}	}
  \end{center}
%\vspace{-0.25in}
\end{table}

\section{Limitation}
Although we provide a comprehensive and in-depth empirical analysis of the strong robustness of large kernel convnets from multiple quantitative and qualitative perspectives, given the black box nature of deep learning, we struggle to provide direct theoretical proofs. Another limitation is due to computational constraints, we conduct ablation of kernel sizes on ImageNet-1K but not ImageNet-21K.

\section{Conclusion}
\label{sec5}
In this study, we thoroughly investigate the robustness of large kernel convnet, validating its strong robustness across six widely-used robustness benchmark datasets. Then we comprehensively analyze the source for its exceptional robustness from both quantitative and qualitative perspectives. Our research and analysis provide novel insights into the source of robustness, potentially promoting large kernel convnet's application and development in the future.

\newpage
\section*{Acknowledgements}
This work is supported in part by the National Key R\&D Program of China (Grant No.2022ZD0116403), the National Natural Science Foundation of China (Grant No. 61721004), and the Strategic Priority Research
Program of Chinese Academy of Sciences (Grant No.
XDA27000000).
\section*{Impact Statement}
This paper presents work whose goal is to advance the field of Machine Learning. There are many potential societal consequences of our work, none of which we feel must be specifically highlighted here.

\bibliography{example_paper}

\begin{thebibliography}{61}
\providecommand{\natexlab}[1]{#1}
\providecommand{\url}[1]{\texttt{#1}}
\expandafter\ifx\csname urlstyle\endcsname\relax
  \providecommand{\doi}[1]{doi: #1}\else
  \providecommand{\doi}{doi: \begingroup \urlstyle{rm}\Url}\fi

\bibitem[Bao et~al.(2021)Bao, Dong, Piao, and Wei]{beit}
Bao, H., Dong, L., Piao, S., and Wei, F.
\newblock Beit: Bert pre-training of image transformers.
\newblock \emph{arXiv preprint arXiv:2106.08254}, 2021.

\bibitem[Bhojanapalli et~al.(2021)Bhojanapalli, Chakrabarti, Glasner, Li, Unterthiner, and Veit]{bhojanapalli2021understanding}
Bhojanapalli, S., Chakrabarti, A., Glasner, D., Li, D., Unterthiner, T., and Veit, A.
\newblock Understanding robustness of transformers for image classification.
\newblock In \emph{Proceedings of the IEEE/CVF international conference on computer vision}, pp.\  10231--10241, 2021.

\bibitem[Carion et~al.(2020)Carion, Massa, Synnaeve, Usunier, Kirillov, and Zagoruyko]{detr}
Carion, N., Massa, F., Synnaeve, G., Usunier, N., Kirillov, A., and Zagoruyko, S.
\newblock End-to-end object detection with transformers.
\newblock In \emph{European conference on computer vision}, pp.\  213--229. Springer, 2020.

\bibitem[Caron et~al.(2021)Caron, Touvron, Misra, J{\'e}gou, Mairal, Bojanowski, and Joulin]{dino}
Caron, M., Touvron, H., Misra, I., J{\'e}gou, H., Mairal, J., Bojanowski, P., and Joulin, A.
\newblock Emerging properties in self-supervised vision transformers.
\newblock In \emph{Proceedings of the IEEE/CVF international conference on computer vision}, pp.\  9650--9660, 2021.

\bibitem[Chen et~al.(2024{\natexlab{a}})Chen, Chu, Ren, Zhao, and Huang]{pelk}
Chen, H., Chu, X., Ren, Y., Zhao, X., and Huang, K.
\newblock Pelk: Parameter-efficient large kernel convnets with peripheral convolution.
\newblock \emph{arXiv preprint arXiv:2403.07589}, 2024{\natexlab{a}}.

\bibitem[Chen et~al.(2024{\natexlab{b}})Chen, Kong, Zhang, Zhao, and Huang]{ddae}
Chen, H., Kong, X., Zhang, X., Zhao, X., and Huang, K.
\newblock Ddae: Towards deep dynamic vision bert pretraining.
\newblock In \emph{Proceedings of the AAAI Conference on Artificial Intelligence}, volume~38, pp.\  1037--1045, 2024{\natexlab{b}}.

\bibitem[Chen et~al.(2023)Chen, Liu, Zhang, Qi, and Jia]{largekernel3d}
Chen, Y., Liu, J., Zhang, X., Qi, X., and Jia, J.
\newblock Largekernel3d: Scaling up kernels in 3d sparse cnns.
\newblock In \emph{Proceedings of the IEEE/CVF Conference on Computer Vision and Pattern Recognition}, pp.\  13488--13498, 2023.

\bibitem[Cheng et~al.(2021)Cheng, Schwing, and Kirillov]{maskformer}
Cheng, B., Schwing, A., and Kirillov, A.
\newblock Per-pixel classification is not all you need for semantic segmentation.
\newblock \emph{Advances in Neural Information Processing Systems}, 34:\penalty0 17864--17875, 2021.

\bibitem[Cheng et~al.(2022)Cheng, Misra, Schwing, Kirillov, and Girdhar]{mask2former}
Cheng, B., Misra, I., Schwing, A.~G., Kirillov, A., and Girdhar, R.
\newblock Masked-attention mask transformer for universal image segmentation.
\newblock In \emph{Proceedings of the IEEE/CVF conference on computer vision and pattern recognition}, pp.\  1290--1299, 2022.

\bibitem[Dai et~al.(2021)Dai, Chen, Yang, Zhang, Yuan, and Zhang]{dynamic_detr}
Dai, X., Chen, Y., Yang, J., Zhang, P., Yuan, L., and Zhang, L.
\newblock Dynamic detr: End-to-end object detection with dynamic attention.
\newblock In \emph{Proceedings of the IEEE/CVF International Conference on Computer Vision}, pp.\  2988--2997, 2021.

\bibitem[Deng et~al.(2009)Deng, Dong, Socher, Li, Li, and Fei-Fei]{imagenet}
Deng, J., Dong, W., Socher, R., Li, L.-J., Li, K., and Fei-Fei, L.
\newblock Imagenet: A large-scale hierarchical image database.
\newblock In \emph{2009 IEEE conference on computer vision and pattern recognition}, pp.\  248--255. Ieee, 2009.

\bibitem[Ding et~al.(2022)Ding, Zhang, Han, and Ding]{replk}
Ding, X., Zhang, X., Han, J., and Ding, G.
\newblock Scaling up your kernels to 31x31: Revisiting large kernel design in cnns.
\newblock In \emph{Proceedings of the IEEE/CVF conference on computer vision and pattern recognition}, pp.\  11963--11975, 2022.

\bibitem[Ding et~al.(2023)Ding, Zhang, Ge, Zhao, Song, Yue, and Shan]{unireplknet}
Ding, X., Zhang, Y., Ge, Y., Zhao, S., Song, L., Yue, X., and Shan, Y.
\newblock Unireplknet: A universal perception large-kernel convnet for audio, video, point cloud, time-series and image recognition.
\newblock \emph{arXiv preprint arXiv:2311.15599}, 2023.

\bibitem[Dosovitskiy et~al.(2020)Dosovitskiy, Beyer, Kolesnikov, Weissenborn, Zhai, Unterthiner, Dehghani, Minderer, Heigold, Gelly, et~al.]{vit}
Dosovitskiy, A., Beyer, L., Kolesnikov, A., Weissenborn, D., Zhai, X., Unterthiner, T., Dehghani, M., Minderer, M., Heigold, G., Gelly, S., et~al.
\newblock An image is worth 16x16 words: Transformers for image recognition at scale.
\newblock \emph{arXiv preprint arXiv:2010.11929}, 2020.

\bibitem[Goodfellow et~al.(2014)Goodfellow, Shlens, and Szegedy]{fgsm}
Goodfellow, I.~J., Shlens, J., and Szegedy, C.
\newblock Explaining and harnessing adversarial examples.
\newblock \emph{arXiv preprint arXiv:1412.6572}, 2014.

\bibitem[Greff et~al.(2016)Greff, Srivastava, and Schmidhuber]{greff2016highway}
Greff, K., Srivastava, R.~K., and Schmidhuber, J.
\newblock Highway and residual networks learn unrolled iterative estimation.
\newblock \emph{arXiv preprint arXiv:1612.07771}, 2016.

\bibitem[Guo et~al.(2023)Guo, Stutz, and Schiele]{vits_robust_6}
Guo, Y., Stutz, D., and Schiele, B.
\newblock Robustifying token attention for vision transformers.
\newblock In \emph{Proceedings of the IEEE/CVF International Conference on Computer Vision}, pp.\  17557--17568, 2023.

\bibitem[He et~al.(2016)He, Zhang, Ren, and Sun]{resnet}
He, K., Zhang, X., Ren, S., and Sun, J.
\newblock Deep residual learning for image recognition.
\newblock In \emph{Proceedings of the IEEE conference on computer vision and pattern recognition}, pp.\  770--778, 2016.

\bibitem[He et~al.(2022)He, Chen, Xie, Li, Doll{\'a}r, and Girshick]{mae}
He, K., Chen, X., Xie, S., Li, Y., Doll{\'a}r, P., and Girshick, R.
\newblock Masked autoencoders are scalable vision learners.
\newblock In \emph{Proceedings of the IEEE/CVF conference on computer vision and pattern recognition}, pp.\  16000--16009, 2022.

\bibitem[Hendrycks \& Dietterich(2019)Hendrycks and Dietterich]{imagenet_cp}
Hendrycks, D. and Dietterich, T.
\newblock Benchmarking neural network robustness to common corruptions and perturbations.
\newblock \emph{arXiv preprint arXiv:1903.12261}, 2019.

\bibitem[Hendrycks et~al.(2020)Hendrycks, Mu, Cubuk, Zoph, Gilmer, and Lakshminarayanan]{augmix}
Hendrycks, D., Mu, N., Cubuk, E.~D., Zoph, B., Gilmer, J., and Lakshminarayanan, B.
\newblock Augmix: A simple method to improve robustness and uncertainty under data shift.
\newblock In \emph{International conference on learning representations}, volume~1, pp.\ ~5, 2020.

\bibitem[Hendrycks et~al.(2021{\natexlab{a}})Hendrycks, Basart, Mu, Kadavath, Wang, Dorundo, Desai, Zhu, Parajuli, Guo, et~al.]{imagenet_r}
Hendrycks, D., Basart, S., Mu, N., Kadavath, S., Wang, F., Dorundo, E., Desai, R., Zhu, T., Parajuli, S., Guo, M., et~al.
\newblock The many faces of robustness: A critical analysis of out-of-distribution generalization.
\newblock In \emph{Proceedings of the IEEE/CVF International Conference on Computer Vision}, pp.\  8340--8349, 2021{\natexlab{a}}.

\bibitem[Hendrycks et~al.(2021{\natexlab{b}})Hendrycks, Zhao, Basart, Steinhardt, and Song]{imagenet_ao}
Hendrycks, D., Zhao, K., Basart, S., Steinhardt, J., and Song, D.
\newblock Natural adversarial examples.
\newblock In \emph{Proceedings of the IEEE/CVF Conference on Computer Vision and Pattern Recognition}, pp.\  15262--15271, 2021{\natexlab{b}}.

\bibitem[Huang et~al.(2023)Huang, Yin, Zhang, Shen, Fang, Pechenizkiy, Wang, and Liu]{lkteacher}
Huang, T., Yin, L., Zhang, Z., Shen, L., Fang, M., Pechenizkiy, M., Wang, Z., and Liu, S.
\newblock Are large kernels better teachers than transformers for convnets?
\newblock \emph{arXiv preprint arXiv:2305.19412}, 2023.

\bibitem[Huang \& Kong(2022)Huang and Kong]{taig}
Huang, Y. and Kong, A. W.-K.
\newblock Transferable adversarial attack based on integrated gradients.
\newblock \emph{arXiv preprint arXiv:2205.13152}, 2022.

\bibitem[Kolesnikov et~al.(2020)Kolesnikov, Beyer, Zhai, Puigcerver, Yung, Gelly, and Houlsby]{bit}
Kolesnikov, A., Beyer, L., Zhai, X., Puigcerver, J., Yung, J., Gelly, S., and Houlsby, N.
\newblock Big transfer (bit): General visual representation learning.
\newblock In \emph{Computer Vision--ECCV 2020: 16th European Conference, Glasgow, UK, August 23--28, 2020, Proceedings, Part V 16}, pp.\  491--507. Springer, 2020.

\bibitem[Kong \& Zhang(2023)Kong and Zhang]{occlusion2}
Kong, X. and Zhang, X.
\newblock Understanding masked image modeling via learning occlusion invariant feature.
\newblock In \emph{Proceedings of the IEEE/CVF Conference on Computer Vision and Pattern Recognition}, pp.\  6241--6251, 2023.

\bibitem[Kosmann-Schwarzbach et~al.(2011)Kosmann-Schwarzbach, Schwarzbach, and Kosmann-Schwarzbach]{occlusion1}
Kosmann-Schwarzbach, Y., Schwarzbach, B.~E., and Kosmann-Schwarzbach, Y.
\newblock \emph{The Noether Theorems}.
\newblock Springer, 2011.

\bibitem[Krizhevsky et~al.(2012)Krizhevsky, Sutskever, and Hinton]{alexnet}
Krizhevsky, A., Sutskever, I., and Hinton, G.~E.
\newblock Imagenet classification with deep convolutional neural networks.
\newblock \emph{Advances in neural information processing systems}, 25, 2012.

\bibitem[LeCun et~al.(1998)LeCun, Bottou, Bengio, and Haffner]{lecun1998gradient}
LeCun, Y., Bottou, L., Bengio, Y., and Haffner, P.
\newblock Gradient-based learning applied to document recognition.
\newblock \emph{Proceedings of the IEEE}, 86\penalty0 (11):\penalty0 2278--2324, 1998.

\bibitem[Lin et~al.(2014)Lin, Maire, Belongie, Hays, Perona, Ramanan, Doll{\'a}r, and Zitnick]{coco}
Lin, T.-Y., Maire, M., Belongie, S., Hays, J., Perona, P., Ramanan, D., Doll{\'a}r, P., and Zitnick, C.~L.
\newblock Microsoft coco: Common objects in context.
\newblock In \emph{Computer Vision--ECCV 2014: 13th European Conference, Zurich, Switzerland, September 6-12, 2014, Proceedings, Part V 13}, pp.\  740--755. Springer, 2014.

\bibitem[Liu et~al.(2022{\natexlab{a}})Liu, Chen, Chen, Chen, Xiao, Wu, Pechenizkiy, Mocanu, and Wang]{slak}
Liu, S., Chen, T., Chen, X., Chen, X., Xiao, Q., Wu, B., Pechenizkiy, M., Mocanu, D., and Wang, Z.
\newblock More convnets in the 2020s: Scaling up kernels beyond 51x51 using sparsity.
\newblock \emph{arXiv preprint arXiv:2207.03620}, 2022{\natexlab{a}}.

\bibitem[Liu et~al.(2021)Liu, Lin, Cao, Hu, Wei, Zhang, Lin, and Guo]{swin}
Liu, Z., Lin, Y., Cao, Y., Hu, H., Wei, Y., Zhang, Z., Lin, S., and Guo, B.
\newblock Swin transformer: Hierarchical vision transformer using shifted windows.
\newblock In \emph{Proceedings of the IEEE/CVF international conference on computer vision}, pp.\  10012--10022, 2021.

\bibitem[Liu et~al.(2022{\natexlab{b}})Liu, Mao, Wu, Feichtenhofer, Darrell, and Xie]{convnext}
Liu, Z., Mao, H., Wu, C.-Y., Feichtenhofer, C., Darrell, T., and Xie, S.
\newblock A convnet for the 2020s.
\newblock In \emph{Proceedings of the IEEE/CVF conference on computer vision and pattern recognition}, pp.\  11976--11986, 2022{\natexlab{b}}.

\bibitem[Lu et~al.(2023)Lu, Ding, Liu, Wu, and Wang]{link}
Lu, T., Ding, X., Liu, H., Wu, G., and Wang, L.
\newblock Link: Linear kernel for lidar-based 3d perception.
\newblock In \emph{Proceedings of the IEEE/CVF Conference on Computer Vision and Pattern Recognition}, pp.\  1105--1115, 2023.

\bibitem[Madry et~al.(2017{\natexlab{a}})Madry, Makelov, Schmidt, Tsipras, and Vladu]{madry2017towards}
Madry, A., Makelov, A., Schmidt, L., Tsipras, D., and Vladu, A.
\newblock Towards deep learning models resistant to adversarial attacks.
\newblock \emph{arXiv preprint arXiv:1706.06083}, 2017{\natexlab{a}}.

\bibitem[Madry et~al.(2017{\natexlab{b}})Madry, Makelov, Schmidt, Tsipras, and Vladu]{pgd}
Madry, A., Makelov, A., Schmidt, L., Tsipras, D., and Vladu, A.
\newblock Towards deep learning models resistant to adversarial attacks.
\newblock \emph{arXiv preprint arXiv:1706.06083}, 2017{\natexlab{b}}.

\bibitem[Mahmood et~al.(2021)Mahmood, Mahmood, and Van~Dijk]{mahmood2021robustness}
Mahmood, K., Mahmood, R., and Van~Dijk, M.
\newblock On the robustness of vision transformers to adversarial examples.
\newblock In \emph{Proceedings of the IEEE/CVF International Conference on Computer Vision}, pp.\  7838--7847, 2021.

\bibitem[Mao et~al.(2022)Mao, Qi, Chen, Li, Duan, Ye, He, and Xue]{vits_robust_4}
Mao, X., Qi, G., Chen, Y., Li, X., Duan, R., Ye, S., He, Y., and Xue, H.
\newblock Towards robust vision transformer.
\newblock In \emph{Proceedings of the IEEE/CVF conference on Computer Vision and Pattern Recognition}, pp.\  12042--12051, 2022.

\bibitem[Meng et~al.(2021)Meng, Chen, Fan, Zeng, Li, Yuan, Sun, and Wang]{conditional_detr}
Meng, D., Chen, X., Fan, Z., Zeng, G., Li, H., Yuan, Y., Sun, L., and Wang, J.
\newblock Conditional detr for fast training convergence.
\newblock In \emph{Proceedings of the IEEE/CVF International Conference on Computer Vision}, pp.\  3651--3660, 2021.

\bibitem[Naseer et~al.(2021)Naseer, Ranasinghe, Khan, Hayat, Shahbaz~Khan, and Yang]{naseer2021intriguing}
Naseer, M.~M., Ranasinghe, K., Khan, S.~H., Hayat, M., Shahbaz~Khan, F., and Yang, M.-H.
\newblock Intriguing properties of vision transformers.
\newblock \emph{Advances in Neural Information Processing Systems}, 34:\penalty0 23296--23308, 2021.

\bibitem[Ortiz-Jimenez et~al.(2020)Ortiz-Jimenez, Modas, Moosavi, and Frossard]{hold_me_tight}
Ortiz-Jimenez, G., Modas, A., Moosavi, S.-M., and Frossard, P.
\newblock Hold me tight! influence of discriminative features on deep network boundaries.
\newblock \emph{Advances in Neural Information Processing Systems}, 33:\penalty0 2935--2946, 2020.

\bibitem[Park \& Kim(2022)Park and Kim]{how}
Park, N. and Kim, S.
\newblock How do vision transformers work?
\newblock \emph{arXiv preprint arXiv:2202.06709}, 2022.

\bibitem[Paul \& Chen(2022)Paul and Chen]{paul2022vision}
Paul, S. and Chen, P.-Y.
\newblock Vision transformers are robust learners.
\newblock In \emph{Proceedings of the AAAI conference on Artificial Intelligence}, volume~36, pp.\  2071--2081, 2022.

\bibitem[Peng et~al.(2017)Peng, Zhang, Yu, Luo, and Sun]{gcn}
Peng, C., Zhang, X., Yu, G., Luo, G., and Sun, J.
\newblock Large kernel matters--improve semantic segmentation by global convolutional network.
\newblock In \emph{Proceedings of the IEEE conference on computer vision and pattern recognition}, pp.\  4353--4361, 2017.

\bibitem[Raghu et~al.(2021)Raghu, Unterthiner, Kornblith, Zhang, and Dosovitskiy]{raghu2021vision}
Raghu, M., Unterthiner, T., Kornblith, S., Zhang, C., and Dosovitskiy, A.
\newblock Do vision transformers see like convolutional neural networks?
\newblock \emph{Advances in neural information processing systems}, 34:\penalty0 12116--12128, 2021.

\bibitem[Romero et~al.(2021{\natexlab{a}})Romero, Bruintjes, Tomczak, Bekkers, Hoogendoorn, and van Gemert]{flexconv}
Romero, D.~W., Bruintjes, R.-J., Tomczak, J.~M., Bekkers, E.~J., Hoogendoorn, M., and van Gemert, J.~C.
\newblock Flexconv: Continuous kernel convolutions with differentiable kernel sizes.
\newblock \emph{arXiv preprint arXiv:2110.08059}, 2021{\natexlab{a}}.

\bibitem[Romero et~al.(2021{\natexlab{b}})Romero, Kuzina, Bekkers, Tomczak, and Hoogendoorn]{ckconv}
Romero, D.~W., Kuzina, A., Bekkers, E.~J., Tomczak, J.~M., and Hoogendoorn, M.
\newblock Ckconv: Continuous kernel convolution for sequential data.
\newblock \emph{arXiv preprint arXiv:2102.02611}, 2021{\natexlab{b}}.

\bibitem[Shao et~al.(2021)Shao, Shi, Yi, Chen, and Hsieh]{shao2021adversarial}
Shao, R., Shi, Z., Yi, J., Chen, P.-Y., and Hsieh, C.-J.
\newblock On the adversarial robustness of vision transformers.
\newblock \emph{arXiv preprint arXiv:2103.15670}, 2021.

\bibitem[Simonyan \& Zisserman(2014)Simonyan and Zisserman]{vgg}
Simonyan, K. and Zisserman, A.
\newblock Very deep convolutional networks for large-scale image recognition.
\newblock \emph{arXiv preprint arXiv:1409.1556}, 2014.

\bibitem[Szegedy et~al.(2015)Szegedy, Liu, Jia, Sermanet, Reed, Anguelov, Erhan, Vanhoucke, and Rabinovich]{szegedy2015going}
Szegedy, C., Liu, W., Jia, Y., Sermanet, P., Reed, S., Anguelov, D., Erhan, D., Vanhoucke, V., and Rabinovich, A.
\newblock Going deeper with convolutions.
\newblock In \emph{Proceedings of the IEEE conference on computer vision and pattern recognition}, pp.\  1--9, 2015.

\bibitem[Szegedy et~al.(2016)Szegedy, Vanhoucke, Ioffe, Shlens, and Wojna]{szegedy2016rethinking}
Szegedy, C., Vanhoucke, V., Ioffe, S., Shlens, J., and Wojna, Z.
\newblock Rethinking the inception architecture for computer vision.
\newblock In \emph{Proceedings of the IEEE conference on computer vision and pattern recognition}, pp.\  2818--2826, 2016.

\bibitem[Touvron et~al.(2021)Touvron, Cord, Douze, Massa, Sablayrolles, and J{\'e}gou]{touvron2021training}
Touvron, H., Cord, M., Douze, M., Massa, F., Sablayrolles, A., and J{\'e}gou, H.
\newblock Training data-efficient image transformers \& distillation through attention.
\newblock In \emph{International conference on machine learning}, pp.\  10347--10357. PMLR, 2021.

\bibitem[Trockman \& Kolter(2022)Trockman and Kolter]{convmixer}
Trockman, A. and Kolter, J.~Z.
\newblock Patches are all you need?
\newblock \emph{arXiv preprint arXiv:2201.09792}, 2022.

\bibitem[Veit et~al.(2016)Veit, Wilber, and Belongie]{veit2016residual}
Veit, A., Wilber, M.~J., and Belongie, S.
\newblock Residual networks behave like ensembles of relatively shallow networks.
\newblock \emph{Advances in neural information processing systems}, 29, 2016.

\bibitem[Wang et~al.(2021)Wang, Xie, Li, Fan, Song, Liang, Lu, Luo, and Shao]{pvt}
Wang, W., Xie, E., Li, X., Fan, D.-P., Song, K., Liang, D., Lu, T., Luo, P., and Shao, L.
\newblock Pyramid vision transformer: A versatile backbone for dense prediction without convolutions.
\newblock In \emph{Proceedings of the IEEE/CVF international conference on computer vision}, pp.\  568--578, 2021.

\bibitem[Woo et~al.(2023)Woo, Debnath, Hu, Chen, Liu, Kweon, and Xie]{convnext_v2}
Woo, S., Debnath, S., Hu, R., Chen, X., Liu, Z., Kweon, I.~S., and Xie, S.
\newblock Convnext v2: Co-designing and scaling convnets with masked autoencoders.
\newblock In \emph{Proceedings of the IEEE/CVF Conference on Computer Vision and Pattern Recognition}, pp.\  16133--16142, 2023.

\bibitem[Xiao et~al.(2020)Xiao, Engstrom, Ilyas, and Madry]{imagenet_9}
Xiao, K., Engstrom, L., Ilyas, A., and Madry, A.
\newblock Noise or signal: The role of image backgrounds in object recognition.
\newblock \emph{arXiv preprint arXiv:2006.09994}, 2020.

\bibitem[Zhai et~al.(2019)Zhai, Puigcerver, Kolesnikov, Ruyssen, Riquelme, Lucic, Djolonga, Pinto, Neumann, Dosovitskiy, et~al.]{vtab}
Zhai, X., Puigcerver, J., Kolesnikov, A., Ruyssen, P., Riquelme, C., Lucic, M., Djolonga, J., Pinto, A.~S., Neumann, M., Dosovitskiy, A., et~al.
\newblock A large-scale study of representation learning with the visual task adaptation benchmark.
\newblock \emph{arXiv preprint arXiv:1910.04867}, 2019.

\bibitem[Zhou et~al.(2022)Zhou, Yu, Xie, Xiao, Anandkumar, Feng, and Alvarez]{vits_robust_5}
Zhou, D., Yu, Z., Xie, E., Xiao, C., Anandkumar, A., Feng, J., and Alvarez, J.~M.
\newblock Understanding the robustness in vision transformers.
\newblock In \emph{International Conference on Machine Learning}, pp.\  27378--27394. PMLR, 2022.

\bibitem[Zhu et~al.(2020)Zhu, Su, Lu, Li, Wang, and Dai]{deformable_detr}
Zhu, X., Su, W., Lu, L., Li, B., Wang, X., and Dai, J.
\newblock Deformable detr: Deformable transformers for end-to-end object detection.
\newblock \emph{arXiv preprint arXiv:2010.04159}, 2020.

\end{thebibliography}
\bibliographystyle{icml2024}

%%%%%%%%%%%%%%%%%%%%%%%%%%%%%%%%%%%%%%%%%%%%%%%%%%%%%%%%%%%%%%%%%%%%%%%%%%%%%%%
%%%%%%%%%%%%%%%%%%%%%%%%%%%%%%%%%%%%%%%%%%%%%%%%%%%%%%%%%%%%%%%%%%%%%%%%%%%%%%%
% APPENDIX
%%%%%%%%%%%%%%%%%%%%%%%%%%%%%%%%%%%%%%%%%%%%%%%%%%%%%%%%%%%%%%%%%%%%%%%%%%%%%%%
%%%%%%%%%%%%%%%%%%%%%%%%%%%%%%%%%%%%%%%%%%%%%%%%%%%%%%%%%%%%%%%%%%%%%%%%%%%%%%%
\newpage
\appendix
\onecolumn

\section{Details of Robustness Datasets}\label{appendix_a}

\textbf{ImageNet-A}~\cite{imagenet_ao} is a dataset of real-world adversarially filtered images that fool current ImageNet classifiers. Specifically, it comprises a 200-class subset of ImageNet-1K’s 1000 classes, covering most broad categories. These images, which should be correctly classified, are instead misclassified by a ResNet-50 with high confidence into incorrect categories. They cause consistent classification mistakes across various models due to scene complications encountered in the long tail of scene configurations and by exploiting classifier blind spots.

\textbf{ImageNet-C}~\cite{imagenet_cp} comprises 15 categories of algorithmically generated corruptions and an additional four general corruption types, resulting in a total of 19 corruption categories.
Each corruption type has five severity levels, ranging from negligible to pulverizing. This range allows the benchmark to provide a comprehensive assessment of each corruption type.
These corruptions are all applied to the ImageNet validation images, so the total data volume for ImageNet-C is $19\times5\times50{\rm k}=4750{\rm k}$ images.

\textbf{ImageNet-O}~\cite{imagenet_ao} is a dataset of adversarially filtered examples for ImageNet out-of-distribution detectors. It comprises of 200 categories from ImageNet-22K that are not included in ImageNet-1K. It contains \textit{anomalies} of unforeseen classes for which a robust model is expected to output low-confidence predictions.

\textbf{ImageNet-P}~\cite{imagenet_cp} consists of 10 types of common perturbations. ImageNet-P differs from ImageNet-C in that it generates perturbation sequences from each ImageNet validation image. The perturbations are subtly nuanced, affecting a smaller number of pixels within the images. To offset the increase in dataset size and evaluation time caused by each sequence containing over 30 frames, it contains only 10 common perturbations. 

\textbf{ImageNet-R}~\cite{imagenet_r} contains various artistic renditions of 200 classes from the ImageNet-1K dataset. Contrary to the original ImageNet, which discouraged such images as the annotators were instructed to collect "photos only, no paintings, no drawings, etc.", ImageNet-R adopts the opposite approach. Aiming at verifying the robustness of vision networks under semantic shifts under different domains. 

\textbf{ImageNet-9}~\cite{imagenet_9} helps disentangle the impacts of foreground and background signals on classification. Human vision exhibits a high degree of robustness to background changes, maintaining consistent decisions as long as the foreground remains the same. However, for most vision models, alterations in the background can significantly impact the model's output and accuracy. Hence, we delve further into the robustness towards background changes.

\begin{table}[h]
%\vspace{-5.5mm}
	\begin{center}
  \caption{Summary of the studied robustness benchmark datasets  on their objectives and corresponding venues.}
   \label{table-dataset}
		\small
		\scalebox{1.0}{\begin{tabular}{l|c|c}
 \toprule
 Dataset & Objective & Venue\\
 \midrule
 ImageNet-A & natural adversarial & CVPR$'21$\\
 ImageNet-C & common corruptions  & ICLR$'19$ \\
 ImageNet-O & out-of-domain distribution & CVPR$'21$ \\
 ImageNet-P & common perturbations  & ICLR$'19$ \\
 ImageNet-R &  semantic shifts&ICCV$'21$  \\
 ImageNet-9 &  background dependency&ICLR$'21$  \\
\bottomrule
 \end{tabular}}
	\end{center}
%\vspace{-3.5mm}
\end{table}

%\section{Stable Feature Map Variance}\label{appendix_b}

\end{document}